\documentclass[11pt]{article}
\usepackage{silence}
\usepackage[final]{acl}

\usepackage{times}
\usepackage{latexsym}

\usepackage[T1]{fontenc}

\usepackage[utf8]{inputenc}

\usepackage{microtype}

\usepackage{inconsolata}

\usepackage{graphicx}
\usepackage{subcaption} %
\usepackage{booktabs} %
\usepackage{makecell} %
\usepackage{adjustbox}%
\usepackage{amsmath}
\usepackage{makecell}
\usepackage{textcomp} 
\usepackage{bm}
\usepackage{booktabs}
\usepackage{tabularx}
\usepackage{array}

\usepackage[utf8]{inputenc}
\usepackage{xcolor}
\usepackage{listings}

\definecolor{codebg}{RGB}{250, 250, 245} 
\definecolor{commentgray}{RGB}{120, 120, 120}

\DeclareFontShape{TS1}{cmtt}{m}{it}{<->ssub * cmtt/m/n}{}
\title{AlphaContext: An Evolutionary Tree-based Psychometric Context Generator for Creativity Assessment}

\author{
\textbf{Yixuan Wang$^{1}$}\thanks{Equal Contribution} \quad
\textbf{Yue Huang$^{1}$}\footnotemark[1] \quad
\textbf{Hong Qian$^{1,2}$}\thanks{Corresponding Author} \quad
\textbf{Yunzhao Wei$^1$} \quad
\textbf{Yifei Ding$^1$} \quad
\\
\textbf{Wenkai Wang$^1$} \quad
\textbf{Zhi Liu$^{1,2}$} \quad
\textbf{Zhongjing Huang$^1$} \quad
\textbf{Aimin Zhou$^{1,2}$} \quad
\textbf{Jiajun Guo$^1$}
\\[4pt]
$^1$East China Normal University, Shanghai, China
\\
$^2$Shanghai Innovation Institute, Shanghai, China
\\[4pt]
\texttt{\{yxwang, yhuang\}@stu.ecnu.edu.cn, hqian@cs.ecnu.edu.cn}
}

\begin{document}
\maketitle
\begin{abstract}
Creativity has become a core competence in the era of LLMs and human–AI collaboration, underpinning innovation in real-world problem solving. Crucially, the systematic improvement of creativity necessitates scientifically valid assessment instruments. Psychometric research recognizes context-based assessment as an effective way to measure creative thinking. However, high-quality expert-designed contexts remain scarce. Existing LLM-based generators often struggle with insufficient assessment cues, weak narrative coherence, limited stylistic diversity, and poor support for creative thinking. To address these challenges, we propose AlphaContext, an evolutionary tree-based psychometric context generator for creativity assessment. First, the HyperTree Outline Planner formalizes expert-designed outlining as a rule-guided hypertree and performs top-down hierarchical planning. The MCTS-based Context Generator fills the outline via MCTS to balance global structure and local quality. Then, the Evolutionary Context Optimizer evolves contexts with MAP-Elites by repeatedly updating niche elites to jointly improve diversity and quality. Finally, the Assessment-Guided Evolution Refiner simulates virtual participants with diverse styles and recycles weak contexts for further evolution. Experiments show that AlphaContext yields an average improvement of 8\% over competitive methods across 6 quality metrics. 
\end{abstract}

\section{Introduction}
Creativity is typically defined as the ability to generate novel and appropriate ideas~\cite{creativity1}, and it is a crucial skill that drives social innovation and scientific discovery~\cite{creativity2}. As AI increasingly takes over routine tasks, creativity is becoming an even more important driver of original contributions and transformative breakthroughs~\cite{creativity-ai}.

\begin{figure}[t]
  \centering
    \includegraphics[width=\linewidth]{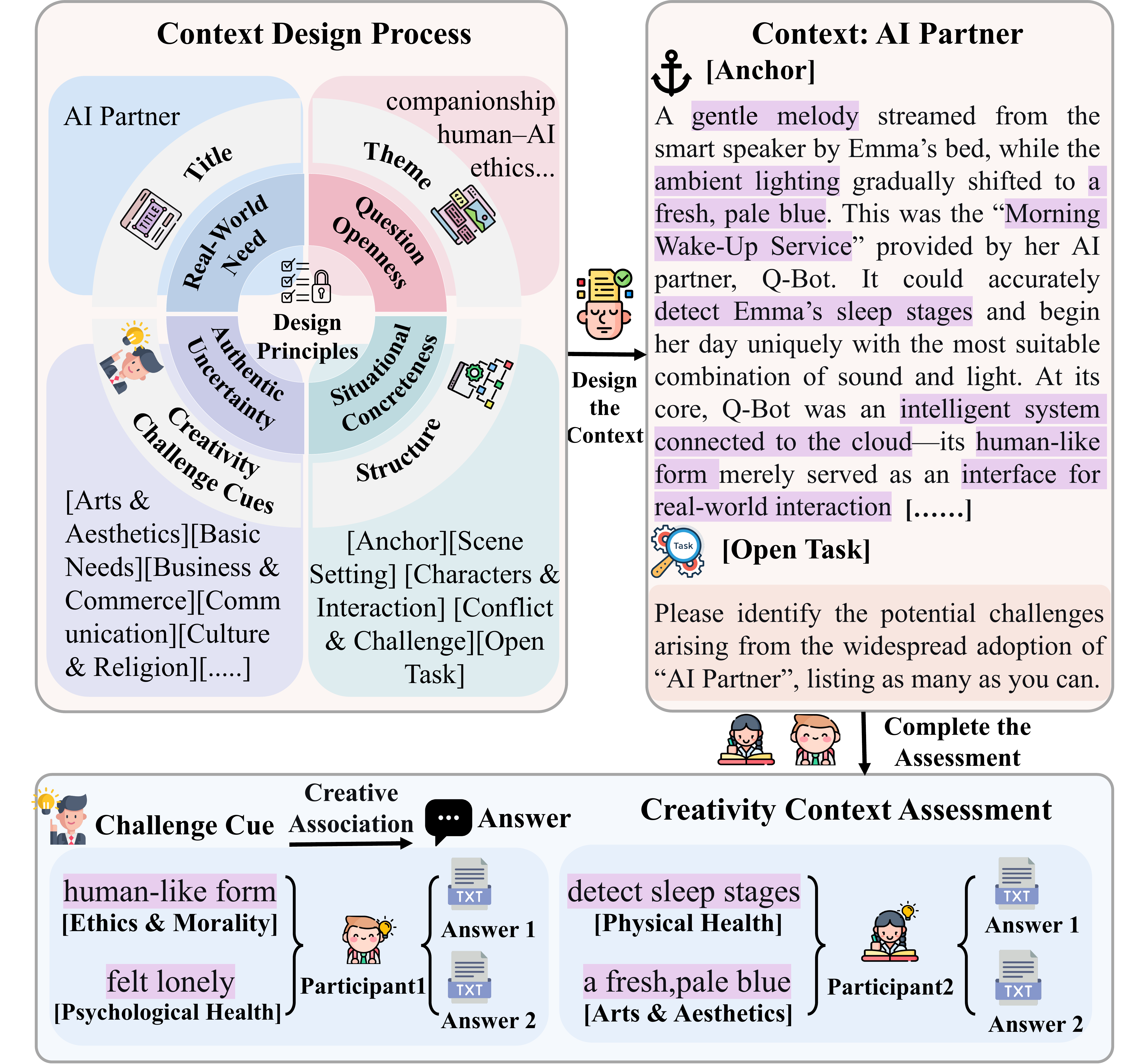}
    \caption{Workflow of creativity context assessment. Experts first design an ``AI Partner'' psychometric context containing implicit challenge cues. Participants then complete an open-ended task to identify potential challenges based on the context in their responses.}
    \label{fig:intro}
\end{figure}

Given this growing importance, scientifically assessing creativity has become a key issue in psychometrics and intelligent education~\cite{item_generator,SAPD,guo}. In creativity assessment, context-based paradigms have been widely adopted~\cite{creativity-ass}. Sternberg's triarchic theory of intelligence emphasizes that creative thinking arises when faced with novel situations~\cite{sternberg}. Therefore, future-oriented contexts, due to their inherent high uncertainty and broad imaginative space, are considered ideal stimuli for eliciting creativity thinking~\cite{future-context}, as shown in Figure~\ref{fig:intro}. The Future Problem Solving Program (FPSP)~\cite{fpsp3} provides authoritative evidence that well-designed future contexts can reliably elicit creative thinking~\cite{fpsp}. Therefore, generating high-quality contexts is essential for developing valid and reliable creativity assessments for humans. However, current practice faces a significant bottleneck in productivity. High-quality creativity assessment contexts still rely on expert craftsmanship. 
 
In recent years, rapid advances in LLMs have substantially improved the generation of stories and dialogues~\cite{storyteller,SS-GEN,drama}, making the automated construction of psychometric contexts for the assessment of creativity increasingly plausible. Although much prior work has examined the creativity of LLMs, there has been far less research on whether LLMs can generate valid creativity assessment contexts for humans. However, psychometric contexts differ from general narratives, and both directly influencing LLMs and reusing story-generation frameworks still face two key challenges. \textbf{\textit{The first key challenge is to simultaneously enforce implicit assessment cues and global narrative coherence}}. Psychometric cues for creative thinking are often embedded implicitly in textual details. Existing methods struggle to precisely control the alignment between cues and themes~\cite{longwriter}, thus failing to satisfy the fine-grained constraints required for psychometric content design and narrative structuring. \textbf{\textit{The second key challenge is to improve diversity while ensuring both context quality and measurement validity at limited cost}}. For a given theme, future problem contexts require diverse types and styles to adapt to different assessment populations~\cite{alphaevolve}, yet increasing diversity typically raises generation and refinement costs. Moreover, creativity assessment contexts require reliable quality and elicitation validity. Traditional expert workflows as shown in Figure~\ref{fig:intro}, rely on expensive human studies and iterative rework~\cite{CollabLLM}, while current methods lack efficient validation and optimization loops. Our work has implications for employing LLMs to automatically generate valid and reliable creativity assessment contexts for humans and AI.

To address these two challenges, this paper proposes AlphaContext, an evolutionary tree-based psychometric context generator for creativity assessment. To tackle the first challenge, the HyperTree Outline Planner formalizes context outlining as a rule-guided hypertree, mapping expert reasoning into a searchable outline space. The MCTS-based Context Generator then performs Monte Carlo Tree Search (MCTS) generation under the outline, balancing global structural coherence and local semantic quality to produce seed contexts. To handle the second challenge, the Evolutionary Context Optimizer conducts evolutionary search with MAP-Elites in a task-specific behavioral space, iteratively expanding stylistic diversity via niche-wise elite updates. Finally, the Assessment-Guided Evolution Refiner simulates virtual participant responses and iteratively refines weak contexts to better elicit creative thinking. Experimental results show that AlphaContext substantially outperforms baselines across multiple evaluation metrics.


\begin{figure*}[t]
  \centering
    \includegraphics[width=\linewidth]{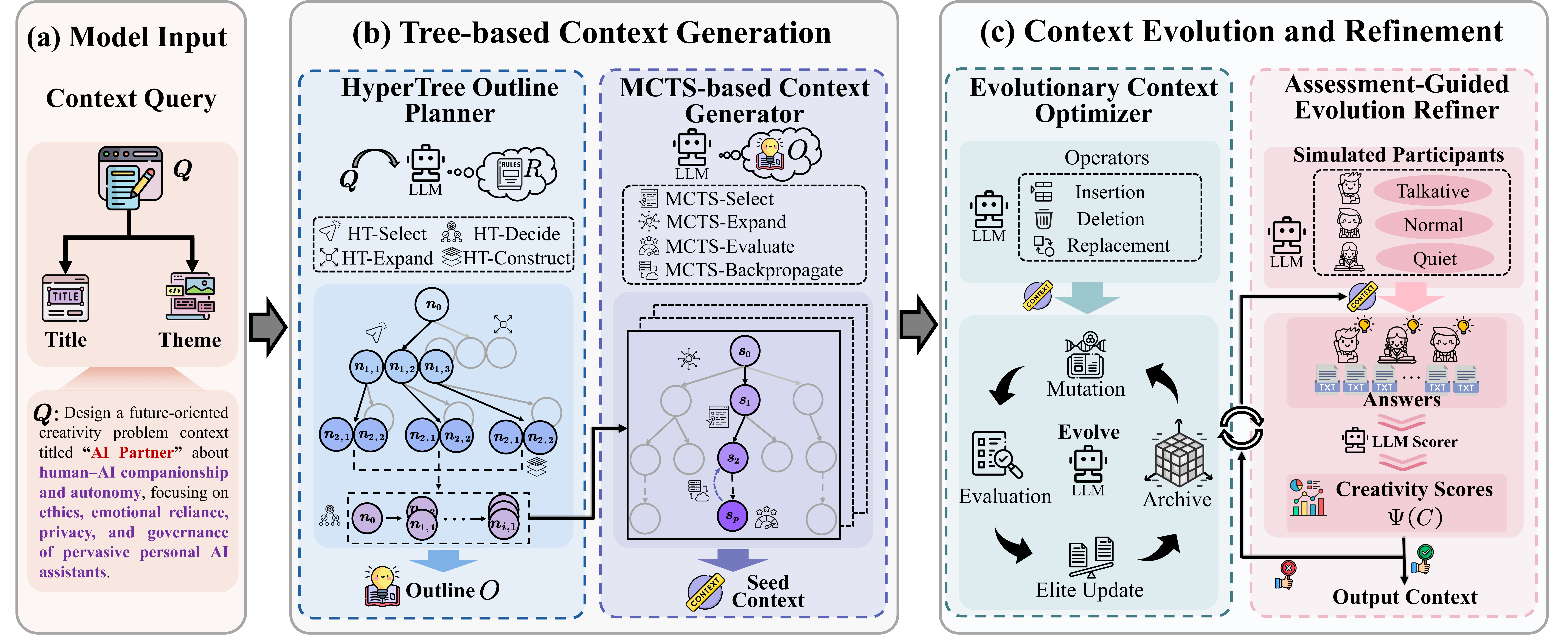}
    \caption{The procedure of the proposed AlphaContext. (a) Given a context query $Q$, (b) the HyperTree Outline Planner and MCTS-based Context Generator generate seed contexts, and (c) the Evolutionary Context Optimizer and Assessment-guided Evolution Refiner improve diversity and quality, yielding assessment-ready contexts.}
    \label{fig:AlphaContext}
\end{figure*}
\section{Related Work}
\subsection{Language Model based Creativity Assessment}
The NLP community has shown growing interest in psychometrically grounded creativity assessment. Luchini et al.~\cite{automated} fine-tuned RoBERTa and GPT-2 to automatically score creativity responses. Since assessment contexts are expert-designed and costly to scale, CPIG investigates using LLMs to automatically generate items for a classic free-response creativity test, examining whether LLMs can generate valid creativity assessments for humans. However, CPIG targets short items and does not address long-form Future Problem contexts, which require discourse-level coherence and implicit assessment cues. 

In parallel, other works evaluate the creativity of LLMs~\cite{SiYH25,Creation-MMbench}. LiveIdeaBench~\cite{LiveIdeaBench} uses single-keyword prompts to assess scientific creative thinking. AidanBench~\cite{aidanbench} measures novelty, non-redundancy, and coherence under open-ended creative questions. Nevertheless, these benchmarks rely on keyword triggers or expert-designed contexts. Moreover, the scarcity of high-quality long-form contexts limits both evaluation protocols and model improvement. 

Overall, the key bottleneck is generating high-quality long-form psychometric contexts that enable valid creativity assessment for humans and also support more comparable evaluation of LLM creativity. To address this gap, we propose AlphaContext, an evolutionary tree-based psychometric context generator for creativity assessment.

\subsection{LLM-based Story Generation}
LLMs have significantly advanced automated narrative generation~\cite{longwriter,drama,LLM_thinking}, enabling the generation of coherent long-form stories. DOC~\cite{DOC} adopts the outline-first strategy and then expands the outline into detailed text. STORYTELLER~\cite{storyteller} introduces a plot node mechanism based on the subject-verb-object (SVO) structure and a dynamic interaction module, further improving narrative coherence and logical consistency. However, most existing story generation methods typically focus on entertainment and fluency, often failing to satisfy the quality and validity requirements of psychometric assessment contexts. Although SS-GEN~\cite{SS-GEN} explores generating psychological social stories for autism interventions, this setting differs fundamentally from creativity assessment settings. Creativity assessment contexts must maintain coherent long-form narratives while implicitly placing assessment cues that elicit creative thinking and support psychometric validity. Consequently, generating high-quality long-form psychometric contexts for creativity assessment remains an open challenge.


\section{Preliminaries}
\noindent\textbf{Creativity Context Generation.}
Given a context design query $Q$ that specifies the title and theme, together with a pre-trained LLM $\pi_\theta$, our goal is to construct an assessment-ready archive $\mathcal{A}=\{C_k\}_{k=1}^{|\mathcal{A}|}$ of contexts for creativity assessment. Each context is represented as $C=(T,O)$, where $O$ is a structured outline, and $T$ is the resulting context text guided by $O$. We adopt a plan-generate-evolve pipeline. The planner produces $O=\pi_\theta(\Phi(Q))$ under predefined instructions $\Phi$. The generator realizes $T$ guided by the outline $O$. The evolve stage iteratively refines and diversifies candidates by updating $\mathcal{A}$.

\noindent\textbf{HyperTree Structure.}
In conventional tree-based planning, each edge links a parent node to a single child node. By contrast, a HyperTree introduces directed hyperedges, where a parent node connects to a set of child nodes via one edge, enabling hierarchical divide-and-conquer by jointly organizing discourse structure and assessment-cue placement for outline planning. Formally, we define a HyperTree as $\mathcal{H}=(N,Q,\mathcal{R})$, where $Q$ denotes the query, $N$ is the node set, and $R$ is a set of expansion rules. Given $Q$, the HyperTree is generated hierarchically according to the rule set $\mathcal{R}$. Compared with ordinary trees, this structure better aligns with expert practices in creativity context design. 


\section{The Proposed AlphaContext}
\textbf{Overview.} As shown in Figure~\ref{fig:AlphaContext}, AlphaContext comprises four modules. Given a query $Q$, the HyperTree Outline Planner places expert outlining as a rule-guided hypertree over a library $\mathcal{R}$. The MCTS-based Context Generator then performs a sentence-level search to fill in the outline. To cover the multi-solution space under the same theme, the Evolutionary Context Optimizer applies MAP-Elites to explore diverse styles in a task-specific behavior space while improving within-niche quality. Finally, the Assessment-Guided Evolution Refiner simulates participant responses and feeds weak contexts back for further evolution.

\subsection{HyperTree Outline Planner}
\label{sec:htp}
Experts plan contexts holistically and refine them hierarchically, motivating a HyperTree representation. This hierarchical organization is supported by cognitive models of narrative memory~\cite{zhong1} and recursive semantic structure in natural language~\cite{zhong2}. We propose the HyperTree Outline Planner to cast outline design as a HyperTree search, where directed hyperedges support hierarchical divide-and-conquer over structure and cue placement. Formally, we define a HyperTree as $\mathcal{H}=(N,Q,\mathcal{R})$. Given $Q$, HyperTree $\mathcal{H}$ is generated hierarchically under $\mathcal{R}$. Each node $n\in N$ corresponds to a structural unit, and each rule $r\in \mathcal{R}$ expands a parent node into a set of child nodes through $r: n_p \mapsto n_c$, where $n_p$ is the parent node and $n_c$ denotes the corresponding child nodes. The HyperTree (HT) planner proceeds in four phases: HT-Select, HT-Expand, HT-Construct, and HT-Decide.

\noindent\textbf{HT-Select.}
Given the current hypertree $\mathcal{H}$, its distinct branches are mapped onto a set of hyperchains $\{L_1,\ldots,L_g\}$, where $g$ denotes the number of hyperchains. To control the search scale, an LLM is used to evaluate and prune candidate hyperchains. Each candidate hyperchain is scored by the LLM and the optimal hyperchains $\mathcal{L}^\ast$ are selected. The divisible nodes are then identified under the rule set $\mathcal{R}$. For each selected hyperchain, we choose its most promising divisible leaf node for expansion using an LLM selector. Thus, this phase consists of two steps: selecting the hyperchains $\mathcal{L}^\ast$ and selecting the divisible leaf node $n_i^\ast$ in each chosen hyperchain $L_i^\ast \in \mathcal{L}^\ast$.

\noindent\textbf{HT-Expand.}
In this phase, given the selected node $n_i^\ast$ in each hyperchain $L_i^\ast$, its applicable expansion rules are retrieved as
$\mathcal{R}(n_i^\ast)=\{\, r\in\mathcal{R}\mid r:n_i^\ast\mapsto n_c \,\}$.
For each rule $r\in\mathcal{R}(n_i^\ast)$, candidate child groups $n_c$ are generated.
Each group is treated as a single branch and is appended to $L_i^\ast$ as the expansion outcome of $n_i^\ast$.

\noindent\textbf{HT-Construct.}
The planner iterates Select and Expand step by step, growing $\mathcal{H}$ from the root node across depth levels over a set of selected hyperchains. Construction stops when no divisible nodes remain or the iteration limit is reached, yielding a hypertree that compactly stores multiple candidate hyperchains.

\noindent\textbf{HT-Decide.}
After constructing $\mathcal{H}$, the LLM globally evaluates the candidate hyperchains and decides the optimal hyperchain as the final outline $O$. This decision jointly considers structural validity for creativity assessment context design and narrative consistency with the input title and theme. More details can be found in Appendix~\ref{sec:apphtp}.

\subsection{MCTS-based Context Generator}
In the generation stage, we cast creativity context writing as a sentence-level decision process guided by an outline $O$. Given an input prompt $x$ and an outline $O$, we build a separate search tree for each discourse section. A context is represented as $C=(T,O)$, where $T=(t_1,\ldots,t_p)$ is the generated sentence sequence. A node at depth $p$ is $s_p=\bigl(t_p,N(s_p),V(s_p),O\bigr)$, where $t_p$ is the current text, $N(s_p)$ is the visit count, and $V(s_p)$ is the estimated value. An LLM policy $\pi_\theta$ proposes the next candidate sentences, and an LLM evaluator provides quality feedback. The generator follows the standard MCTS loop: MCTS-Select, MCTS-Expand, MCTS-Evaluate, and MCTS-Backpropagate.

\noindent \textbf{MCTS-Select.}
Starting from the root $s_0$, the search recursively selects the child with the highest exploration potential according to the UCT score. Specifically, the UCT score of the node $s_p$ is defined as $\mathrm{UCT}(s_p)= V(s_p) + c \sqrt{\frac{\ln N(q)}{N(s_p)}}$. 
Here, $V(s_p)$ denotes the value score of $s_p$, $N(s_p)$ is its visit count, and $N(q)$ is the visit count of its parent node $q$. $c$ is a hyper-parameter that balances exploitation ($V(s_p)$) and exploration (the second term).

\noindent \textbf{MCTS-Expand.}
Given the selected node $s_p$, we expand it by sampling $U$ candidate next sentences from the policy model $\pi_\theta$: $t_{p+1}^{(u)} \sim \pi_\theta(\cdot\!\mid\! x,\, t_{1:p},\, O),\; u=1,\ldots,U$. Here, $t_{1:p}$ denotes the previously generated sentences, enabling parallel exploration of diverse narrative realizations and cue instantiations within $O$. 

\noindent \textbf{MCTS-Evaluate.}
We evaluate each expanded node to assign its node value $V(s_{p+1})$. For long-form creativity context generation, we adopt a dual-horizon valuation mechanism at the evaluation phase to balance reliability and computational cost. Given an expanded child $s_{p+1}$, we first apply multi-aspect immediate scoring with an evaluator:
\begin{equation}
\label{eq6}
V_{\mathrm{imm}}(s_{p+1})=\bar S(s_{p+1})\bigl(1-S_{\mathrm{ha}}(s_{p+1})\bigr).
\end{equation}
Here, $\bar S(s_{p+1})$ is a weighted average of cue alignment $S_{\mathrm{sc}}$, imagery vividness $S_{\mathrm{im}}$, and discourse coherence $S_{\mathrm{co}}$ with $\sum_i \omega_i=1$. $S_{\mathrm{ha}}(s_{p+1})$ measures hallucination risk. To mitigate myopic decisions, when $V_{\mathrm{imm}}(s_{p+1})<\tau$, we sample a short continuation and re-evaluate the concatenated fragment to obtain a more stable value estimate, a lightweight look-ahead for this node. This look-ahead is triggered for low-scoring nodes to reduce myopic errors, while high-confidence nodes directly use the immediate evaluation to save sampling budget. For more details, please refer to our Appendix~\ref{sec:mcts}.

\noindent \textbf{MCTS-Backpropagate.}
The obtained evaluation score $r_e$ is propagated back along the simulated path to all ancestor nodes $s_j$ ($0\le j\le p$), updating the visit counts and value estimates:
\begin{equation}
\label{eq7}
\begin{aligned}
N_{\mathrm{new}}(s_j) &= N_{\mathrm{old}}(s_j) + 1\,, 
\end{aligned}
\end{equation}
\begin{equation}
\label{eq7-2}
\begin{aligned}
V_{\mathrm{new}}(s_j) &=
\frac{V_{\mathrm{old}}(s_j)\,N_{\mathrm{old}}(s_j) + r_e}{N_{\mathrm{new}}(s_j)}\,.
\end{aligned}
\end{equation}
After multiple simulations, the tree concentrates on trajectories that better satisfy the outline, improve coherence, and reduce hallucination risk. We then extract the highest-value root-to-leaf path as a seed context to initialize the evolutionary module.

\subsection{Evolutionary Context Optimizer} 
We introduce a MAP-Elites Evolutionary Context Optimizer initialized with the MCTS seed context. It maintains an elite archive in a style-oriented behavior space, expanding coverage and improving within-niche quality. We next describe the archive, mutation, evaluation, and update rules.

\noindent \textbf{Diversity Archive.}
To characterize stylistic variations in creativity assessment contexts, we map each candidate context $C$ into the behavior space $B$ with a descriptor function $b(\cdot)$:
\begin{equation}
\label{eq8}
b(C)=\bigl[\phi_1(C),\phi_2(C),\phi_3(C)\bigr]\in[0,1]^3\,.
\end{equation}
Here, $\phi_1$ captures proximity scope, measuring the extent to which a context is framed from personal daily-life settings to broader public issues. $\phi_2$ captures knowledge density, reflecting how strongly the narrative is grounded in objective evidence such as data, mechanisms, and causal explanations. $\phi_3$ captures viewpoint diversity, indicating the breadth of stakeholders involved and the need for multi-perspective integration. We uniformly discretize $[0,1]^3$ to form a 3D grid archive, where each cell defines a behavioral niche and stores the current elite context with the highest fitness.

\noindent \textbf{Mutation.}
In natural language space, we implement mutation as a conditional LLM editing policy $\pi_\theta$ that edits a parent elite context $C_p$. At each iteration, we apply an operator set $\Omega=\{\textsc{Insertion},\textsc{Deletion},\textsc{Replacement}\}$ to revise key paragraphs and cue-bearing units, introducing stylistic variation while maintaining assessment-critical content.

\noindent \textbf{Evaluation.}
Given a mutated candidate $C$, we perform both behavior feature evaluation and quality evaluation. Feature evaluation computes $b(C)$ to determine the candidate's niche assignment. Quality evaluation is produced by an LLM-based scorer in terms of narrative coherence $S_{\mathrm{coh}}$, topical relevance $S_{\mathrm{rel}}$, and engagement $S_{\mathrm{eng}}$, and we define fitness $F(C)=\mathrm{Avg}\Bigl(S_{\mathrm{coh}}(C)+S_{\mathrm{rel}}(C)+S_{\mathrm{eng}}(C)\Bigr)$. The scorer outputs three normalized quality scores for each candidate context, and the fitness is defined as their uniform average to avoid introducing extra hyperparameters.

\noindent \textbf{Elite Update.}
After evaluation, we assign context $C$ to the grid niche indexed by $b(C)$. If the niche is empty, $C$ is inserted as the initial elite. Otherwise, let $C^\ast$ denote the current elite in the niche; we replace $C^\ast$ only when $F(C) > F(C^\ast)$.

\subsection{Assessment-Guided Evolution Refiner}
To ensure that generated creativity contexts reliably elicit measurable creative thinking, we propose an Assessment-Guided Evolution Refiner. Concretely, we instantiate an LLM-based participant simulator with a temperature set to $0$ to suppress sampling randomness and stabilize response generation, making contexts comparable in a consistent simulation setting. To improve realism and interpretability, we model response styles as explicit profiles grounded in common psychometric response patterns and enforce each profile via role-conditioned prompting. We consider three profiles—\textit{talkative}, \textit{normal}, and \textit{quiet}. Given a candidate context $C$, the simulator generates a set of responses $\{Y_m\}_{m=1}^M$. The refiner then scores each response using a creativity scorer $f_{\mathrm{cre}}(\cdot)$. We define the assessment effectiveness of a context as the average creativity score across styles $\Psi(C) = \frac{1}{M}\sum_{m=1}^{M} f_{\mathrm{cre}}(Y_m)$. If $\Psi(C)$ exceeds an expert-specified threshold, we treat $C$ as an assessment-ready context. Otherwise, we route $C$ back to the Evolutionary Context Optimizer for further iterative optimization. 

\begin{table*}[htbp]
\centering
\caption{Performance comparison across different methods on the CreaTE dataset. AlphaContext achieves the best results across all seven perspectives. All metrics are presented as positive percentages, where higher values indicate better performance. For each metric, the best-performing model is highlighted in \textbf{bold} and the second is \underline{underlined}.}
\resizebox{\textwidth}{!}{
\begin{tabular}{l|cccccccc}
\toprule
\textbf{Methods}           & \textbf{\makecell[c]{Coherence\\(\textuparrow)\%}} 
& \textbf{\makecell[c]{Relevance\\(\textuparrow)\%}} 
& \textbf{\makecell[c]{Engagement\\(\textuparrow)\%}} 
& \textbf{\makecell[c]{Significance\\(\textuparrow)\%}} 
& \textbf{\makecell[c]{Concreteness\\(\textuparrow)\%}} 
& \textbf{\makecell[c]{Uncertainty\\(\textuparrow)\%}} 
& \textbf{\makecell[c]{Diverse\\Verbs (\textuparrow)\%}} \\
\midrule
DeepSeek-V3.1                  & 50.00  & 50.00  & 50.00  & 50.00  & 50.00  & 50.00  & \underline{94.33}  \\ 
Qwen3-235B-A22B                & 41.50  & 47.91  & 45.07  & 33.87  & 46.55  & 53.33  & 92.09         \\ 
Llama3.3-70B-Instruct          & 27.83  & 26.97  & 33.99  & 27.46  & 30.54  & 20.07  & 90.39         \\        
LongWriter-llama3.1-8b         & 26.60  & 27.46  & 28.63  & 23.40  & 33.62  & 25.99  & 91.18          \\        
LongWriter-glm4-9b             & 32.27  & 31.40  & 31.38  & 36.19  & 32.98  & 31.32  & 88.69          \\ 
GPT-5.1                        & 70.44  & 70.20  & \underline{65.39}   & 50.37  & \underline{71.80} & \underline{68.60}  & 92.88 \\ 
Gemini-3.0-Pro-Preview        & \underline{72.54}   & \underline{75.37}   & 62.56  & 48.40  & 64.16  & 63.30  & 91.81 \\ 
DOC-v2                         & 49.14  & 61.33  & 61.82  & 34.98  & 51.11  & 43.10  & 92.82  \\
CRITICS                        & 51.11  & 61.95  & 61.21  & 37.81  & 54.43  & 42.12  & 92.31 \\ 
SS-GEN                        & 60.22  & 69.69  & 56.40  & \underline{60.10}  & 51.85  & 53.57  & 90.24 \\ 
\midrule
\textbf{AlphaContext}   & \textbf{81.28}     & \textbf{79.06}    & \textbf{79.93}    & \textbf{71.06}                
                        & \textbf{75.49}     & \textbf{80.30}    & \textbf{96.06}                \\
\bottomrule
\end{tabular}
}
\label{tab:exp1}
\end{table*}

\section{Experiments}
This section first describes the CreaTE dataset and details the evaluation metrics used. We then conduct extensive experiments to answer the following research questions. The codes and data are available at \url{https://github.com/yxwang19/AlphaContext}.

\noindent\textbf{Q1:} How does AlphaContext compare to existing methods in generating high-quality creativity contexts across multiple evaluation dimensions?

\noindent\textbf{Q2:} To what extent do the core components contribute to the performance of AlphaContext?

\noindent\textbf{Q3:} How does AlphaContext compare to other methods in terms of textual similarity with expert-designed contexts?

\noindent\textbf{Q4:} Does the LLM-based judge align with human preferences to support reliable evaluation?

\noindent\textbf{Q5:} Does AlphaContext remain effective for creativity assessment in real-world human studies?

\noindent\textbf{Q6:} How does the correlation of AlphaContext with expert-designed assessments compare to that of the strong LLM baseline?

\noindent\textbf{Q7:} What is the computational cost of AlphaContext in generation time and token consumption?


\subsection{Experimental Setup}

\noindent\textbf{Dataset.}
We evaluate AlphaContext on CreaTE. Since general story-generation datasets prioritize narrative completeness and style rather than assessment alignment, we construct \textbf{\textit{CreaTE}}: 203 expert-curated title–theme inputs balancing evaluation cost and domain coverage. Three creativity-psychology experts write each entry and conduct iterative cross-checks. An entry is included only after consensus validation and revision, and all inputs are further screened to remove sensitive information. Details are provided in Appendix~\ref{sec:dataset}.
\begin{table*}[htbp]
\centering
\caption{Ablation study of AlphaContext on multiple evaluation metrics. Details are the same as Table~\ref{tab:exp1}.}
\resizebox{0.9\textwidth}{!}{
\begin{tabular}{l|cccccccc}
\toprule
\textbf{Methods}           & \textbf{\makecell[c]{Coherence\\(\textuparrow)\%}} 
& \textbf{\makecell[c]{Relevance\\(\textuparrow)\%}} 
& \textbf{\makecell[c]{Engagement\\(\textuparrow)\%}} 
& \textbf{\makecell[c]{Significance\\(\textuparrow)\%}} 
& \textbf{\makecell[c]{Concreteness\\(\textuparrow)\%}} 
& \textbf{\makecell[c]{Uncertainty\\(\textuparrow)\%}} 
& \textbf{\makecell[c]{Diverse\\Verbs (\textuparrow)\%}} \\
\midrule
-w/o HOP      & 77.96   & 70.20  & 76.85  & 63.55  & 70.69  & 76.11  & 94.25 \\
-w/o MCG      & 74.38   & 71.80  & 72.17  & 65.76  & 69.09  & 71.92  & 93.79 \\
-w/o ECO      & 75.62   & 70.57  & 71.80  & 64.53  & 68.72  & 70.69  & 93.36  \\
\midrule
\textbf{AlphaContext}   & \textbf{81.28}     & \textbf{79.06}    & \textbf{79.93}    & \textbf{71.06}                
                        & \textbf{75.49}     & \textbf{80.30}    & \textbf{96.06}                \\
\bottomrule
\end{tabular}
}

\label{tab:exp2}
\end{table*}

\begin{table}[t]
\centering
\caption{Text similarity to expert-designed contexts measured by ROUGE-1, ROUGE-L, and BERTScore. Higher is better.}
\small
\scalebox{0.9}{
\begin{tabular}{l|ccc}
\toprule
\textbf{Methods}      & \textbf{\makecell[c]{ROUGE-1\\(\textuparrow)\%}} 
& \textbf{\makecell[c]{ROUGE-L\\(\textuparrow)\%}} 
& \textbf{\makecell[c]{BERTScore\\(\textuparrow)\%}} \\
\midrule
DeepSeek-V3.1                   & 26.22 & 20.53 & 80.94 \\
Qwen3-235B-A22B                 & 20.03 & 16.42 & 80.39 \\
Llama3.3-70B-Instruct           & 20.23 & 16.32 & 80.81 \\
LongWriter-llama3.1-8b          & 19.88 & 16.33 & \underline{81.42} \\
LongWriter-glm4-9b              & 24.42 & 20.34 & 81.33 \\
GPT-5.1                         & 15.25 & 12.46 & 79.28 \\
Gemini-3.0-Pro-Preview         & 22.89 & 18.59 & 80.31 \\
DOC-v2                          & 18.24 & 15.65 & 79.87 \\
CRITICS                         & 17.83 & 15.14 & 79.74 \\
SS-GEN                          & \underline{27.80} & \underline{21.33} & 81.07 \\
\midrule
\textbf{AlphaContext}           & \textbf{30.41} & \textbf{25.48} & \textbf{81.88} \\
\bottomrule
\end{tabular}
}

\label{tab:exp3}
\end{table}

\noindent\textbf{Baselines.} To compare AlphaContext with both strong LLMs and generation frameworks, we consider baselines from three categories. The prompt template is provided in Appendix~\ref{sec:prompt}.

\textbf{(i) General-purpose LLMs.} We consider DeepSeek-V3.1, Qwen3-235B-A22B, Llama3.3-70B-Instruct, GPT-5.1, and Gemini-3.0-Pro-Preview as competitive instruction-following models with strong general reasoning and generation capabilities.
\textbf{(ii) Long-form specialized LLMs.}
LongWriter-llama3.1-8b and LongWriter-glm4-9b~\cite{longwriter} are included as specialized baselines for long-form writing.
\textbf{(iii) Structured generation frameworks.}
DOC-v2~\cite{DOC} combines hierarchical outlining with an adherence controller. CRITICS~\cite{CRITICS} performs critic-guided iterative refinement. SS-GEN~\cite{SS-GEN} applies constraint-driven hierarchical prompting (STARSOW) for structured story generation. We do not compare with CPIG since it generates short test items and is not designed for long-form context generation.

\noindent\textbf{Evaluation Metrics.}
We evaluate each generated creativity context along 7 dimensions. Coherence measures narrative consistency, Relevance measures theme alignment, and Engagement measures how motivating the context is for participants. Our evaluation framework is theoretically grounded in Amabile's Componential Model of Creativity~\cite{amabile1983social,amabile2018creativity}, which emphasizes task motivation, domain-relevant grounding, and creativity-related processes as core foundations of creative performance. We further evaluate Significance~\cite{significance1,significance2} to capture real-world relevance and intrinsic task motivation, Concreteness~\cite{concreteness} to reflect situational specificity that supports feasible ideation, and Uncertainty~\cite{uncertainty1,uncertainty2} to quantify open-endedness that fosters divergent thinking rather than premature closure. 

These psychometric dimensions were iteratively refined through extensive reviews by senior experts in creativity psychology and aligned with established standards from the Future Problem Solving Program (FPSP)~\cite{fpsp3}, which supports strong content validity. Following arena-hard-auto~\cite{Arena}, we use contexts generated by DeepSeek-V3.1 as the reference baseline and obtain quantified scores through pairwise comparisons with other generated contexts. We also report Diverse Verbs~\cite{structure_story} to measure action diversity in the context. The prompt template is provided in Appendix~\ref{sec:eva_prompt}. We also report ROUGE-1, ROUGE-L, and BERTScore to measure similarity to expert-designed contexts. Empirically, our real-world human study further supports construct validity by showing significant positive correlations with standardized creativity measures.

\noindent\textbf{Implementation Details.}
All open-source models are locally deployed and run with vLLM on 8$\times$ NVIDIA H200 GPUs. For DOC-v2, CRITICS, and SS-GEN, DeepSeek-V3.1 is used as the generation engine. Additionally, DeepSeek-V3.1 serves as the evaluator model for all judgments. To mitigate position bias in pairwise judgments, we evaluate each context pair twice with swapped orders, and repeat this procedure for two rounds, resulting in four evaluations per pair. We omit standard deviations in tables since they are consistently small.

\subsection{Experimental Results and Analysis}

\noindent\textbf{Main Performance Evaluation (To Q1).}
We compare AlphaContext with 10 baselines on CreaTE to assess multi-dimensional context quality for creativity assessment. Diverse Verbs is computed automatically, while Coherence, Relevance, Engagement, Significance, Concreteness, and Uncertainty are judged by an LLM. We follow arena-hard-auto~\cite{Arena} and conduct pairwise comparisons against the DeepSeek-V3.1 output as the reference model. For each subjective metric, we report the positive rate over the reference, which is set at 50\% by definition. Table~\ref{tab:exp1} summarizes the results on CreaTE. AlphaContext ranks first on all seven metrics, with the largest gains on Coherence, Engagement, Significance, and Uncertainty, which are key for constructing coherent stimuli that implicitly cue challenges and elicit open-ended creative thinking. Notably, SS-GEN surpasses GPT-5.1 and Gemini-3.0-Pro-Preview on Significance, while AlphaContext further raises it to 71.06\%. This suggests that AlphaContext's planning, search-based generation, and iterative optimization collectively drive consistent improvements across metrics.


\noindent\textbf{Ablation Study (To Q2).}
To quantify module contributions, we build three ablated variants by removing the HyperTree Outline Planner (HOP), the MCTS-based Context Generator (MCG), and the Evolutionary Context Optimizer (ECO). Table~\ref{tab:exp2} shows that removing any module degrades performance. In particular, removing HOP yields a sharp Relevance drop to 70.20\%, indicating that hierarchical planning is critical for keeping assessment cues aligned with the intended theme. Removing MCG lowers Coherence to 74.38\% and also reduces Engagement and Uncertainty, suggesting that MCTS search helps preserve long-range structure and the open-endedness needed to elicit creative thinking. Removing ECO decreases all metrics, most notably Uncertainty, showing that MAP-Elites refinement is important for expanding stylistic coverage while maintaining assessment cues. This decline arises from two complementary roles of the ECO. First, it expands stylistic coverage by maintaining niche-specific elites in a 3D behavior space, which broadens the diversity of generated contexts. Second, it acts as an effective quality filter through the iterative mutation–evaluation–update loop, which polishes consistency and theme alignment beyond raw MCTS seeds. The ablation results thus validate that the MAP-Elites refinement simultaneously enhances stylistic diversity and core quality while preserving assessment cues. Overall, the ablation results show that the three modules make complementary contributions.

\begin{figure}[t]
  \centering

  \begin{subfigure}[b]{0.48\linewidth}
    \centering
    \includegraphics[width=\linewidth]{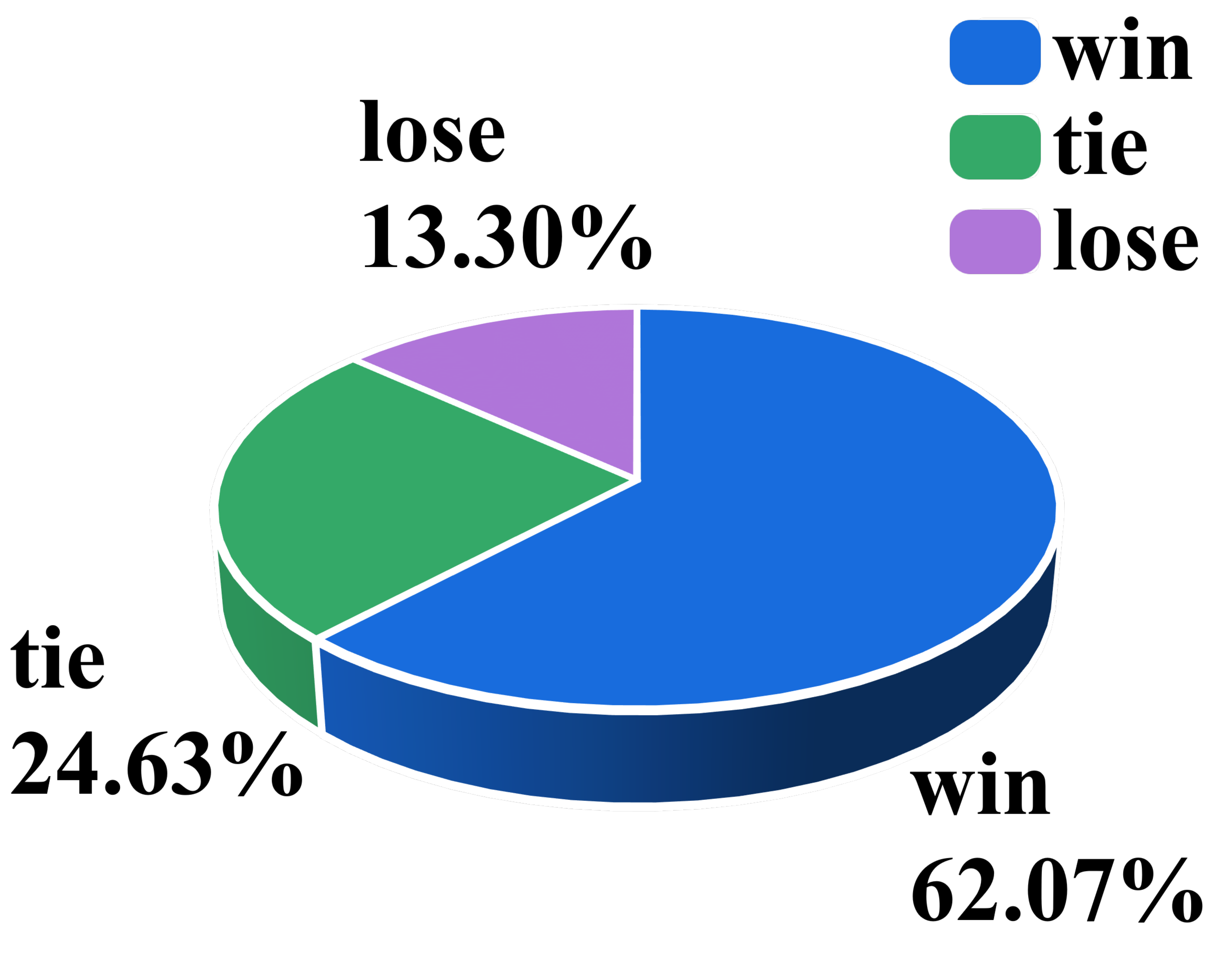}
    \caption{\centering AlphaContext vs.\ GPT-5.1 \\ (Human)}
    \label{fig:pie_gpt_human}
  \end{subfigure}
  \hfill
  \begin{subfigure}[b]{0.48\linewidth}
    \centering
    \includegraphics[width=\linewidth]{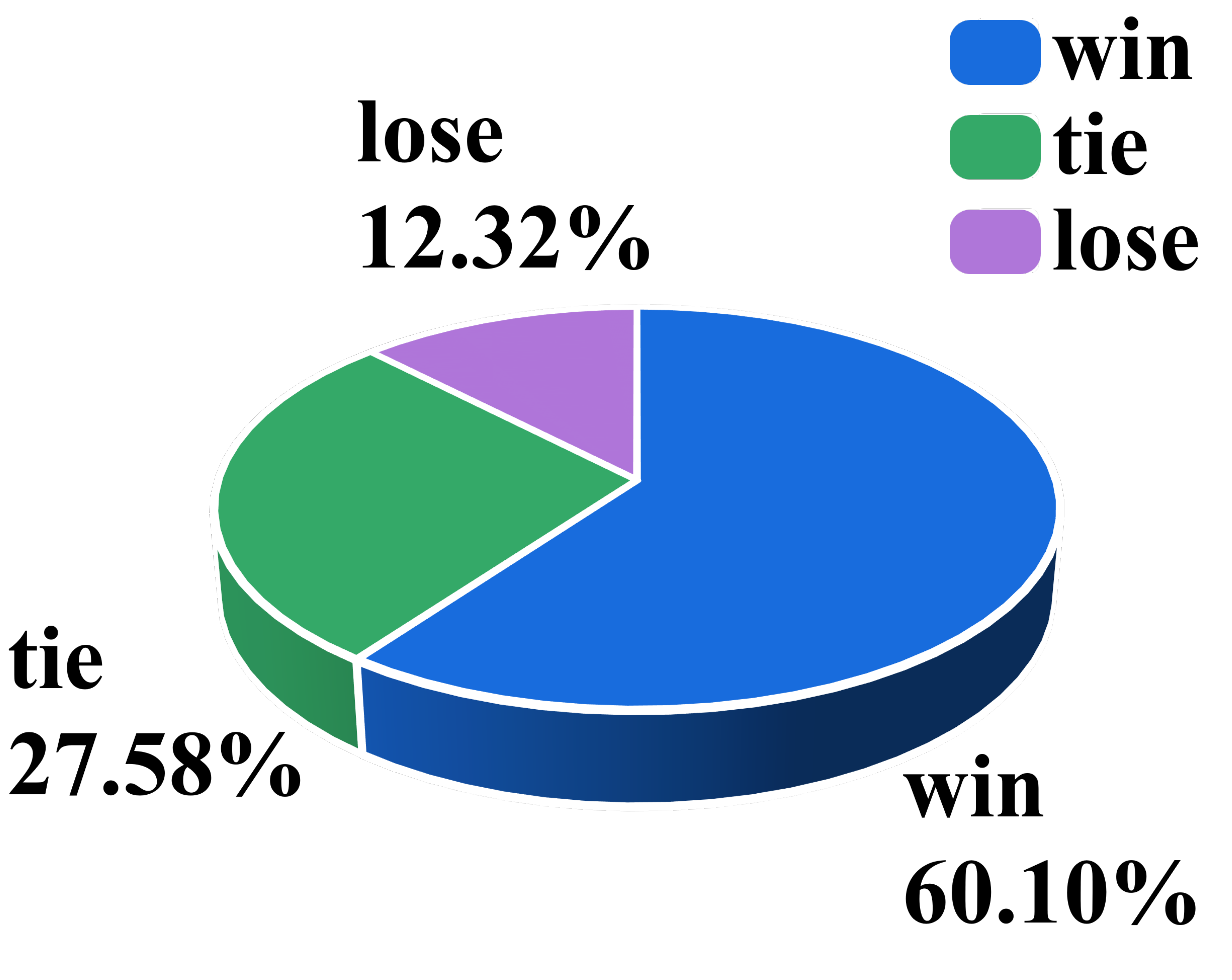}
    \caption{\centering AlphaContext vs.\ GPT-5.1 \\ (DeepSeek-V3.1)}
    \label{fig:pie_gpt_deepseek}
  \end{subfigure}

  \caption{Preference evaluation of AlphaContext vs.\ GPT-5.1 under human and DeepSeek-V3.1 judgments.}
  \label{fig:pie_gpt_pair}
\end{figure}

\begin{figure}[t]
  \centering

  \begin{subfigure}[b]{0.48\linewidth}
    \centering
    \includegraphics[width=\linewidth]{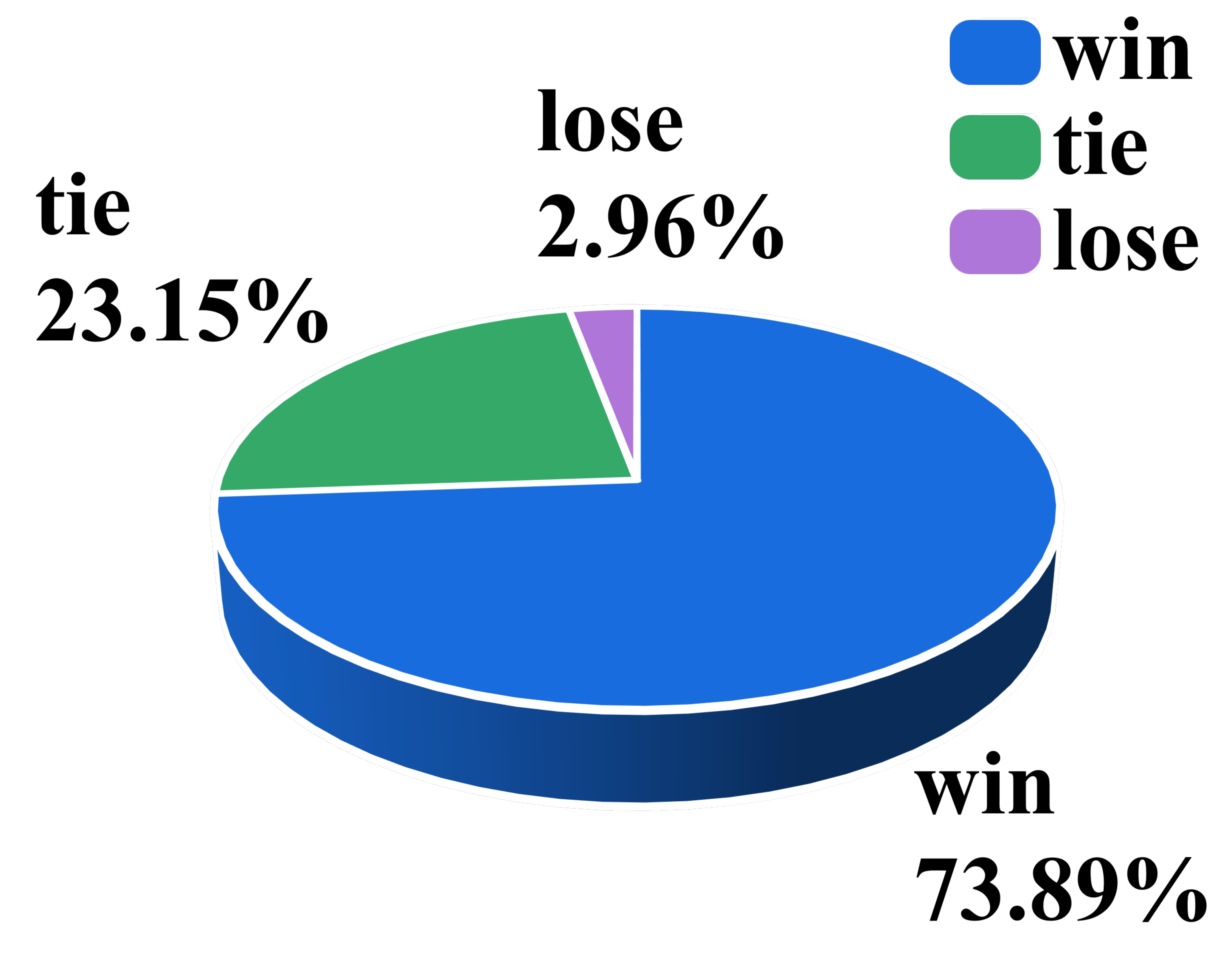}
    \caption{\centering AlphaContext vs.\ Gemini \\ (Human)}
    \label{fig:pie_gemini_human}
  \end{subfigure}
  \hfill
  \begin{subfigure}[b]{0.48\linewidth}
    \centering
    \includegraphics[width=\linewidth]{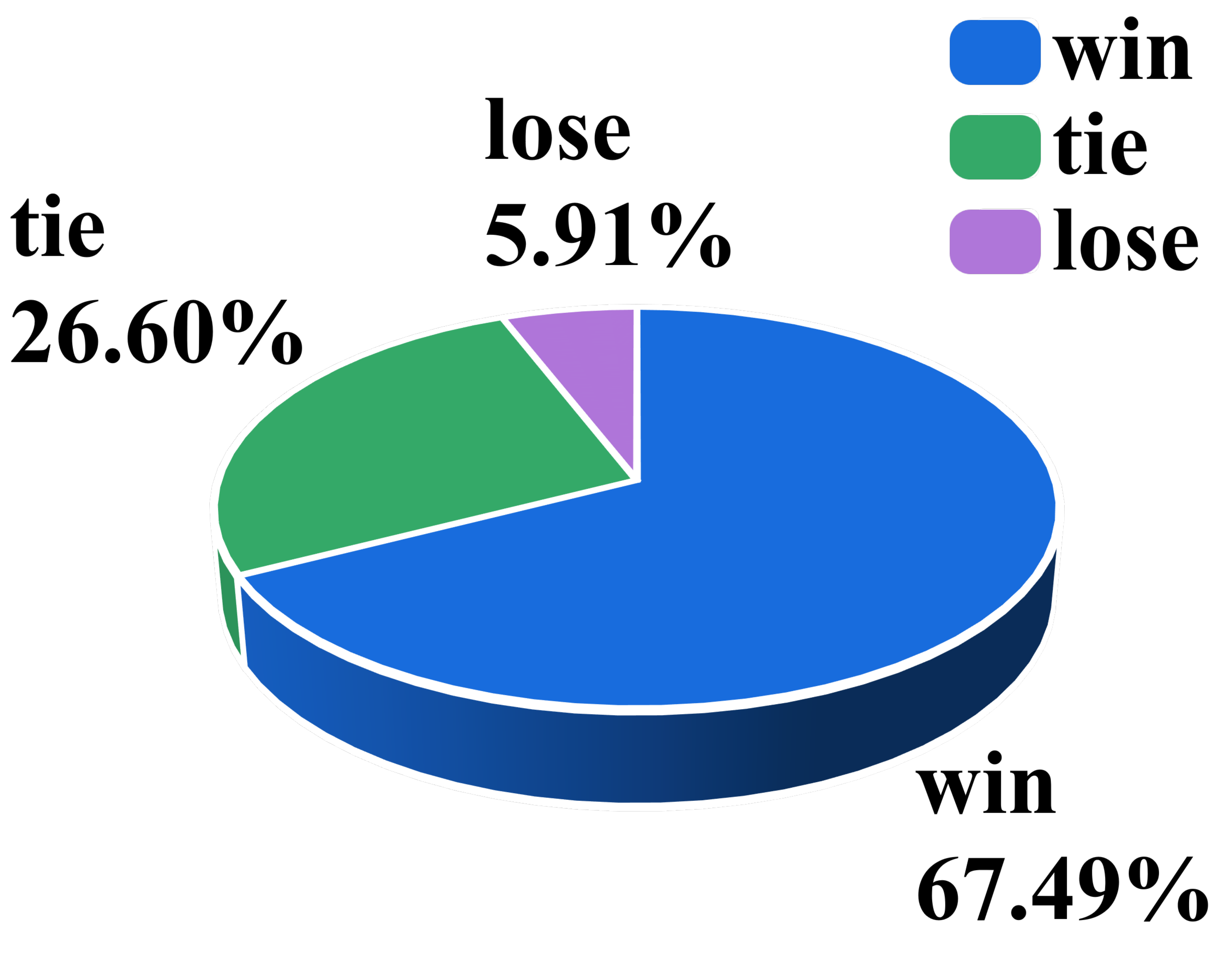}
    \caption{\centering AlphaContext vs.\ Gemini \\ (DeepSeek-V3.1)}
    \label{fig:pie_gemini_deepseek}
  \end{subfigure}

  \caption{Preference evaluation of AlphaContext vs.\ Gemini-3.0-Pro-Preview under human and DeepSeek-V3.1 judgments.}
  \label{fig:pie_gemini_pair}
\end{figure}

\noindent\textbf{Context Similarity Evaluation (To Q3).}
To assess how closely AlphaContext outputs match expert-designed contexts given the same inputs, we use 16 expert contexts that were deployed in real creativity assessments and validated by domain experts as references. We compute ROUGE-1, ROUGE-L, and BERTScore, which reflect lexical overlap, long-span matching, and semantic similarity, respectively. Table~\ref{tab:exp3} shows that AlphaContext achieves the best results in all three metrics. It reaches 30.41\% ROUGE-1 and 25.48\% ROUGE-L, outperforming SS-GEN by 2.61\% and 4.15\%, and also surpasses strong LLM baselines in BERTScore. Overall, AlphaContext matches the most closely expert-designed contexts, aligning with its goal of generating psychometrically grounded creativity assessment materials. For more details, please refer to our Appendix~\ref{sec:gemini}.

\noindent\textbf{Preference Evaluation (To Q4).}
To examine whether the LLM judge aligns with human preferences, we compare AlphaContext with strong baselines (GPT-5.1 and Gemini-3.0-Pro-Preview) via pairwise preference judgments from both human evaluators and DeepSeek-V3.1. To mitigate position bias, each pair is judged twice with swapped orders, and we report win, tie, and lose rates, where win indicates a preference for AlphaContext. Figure~\ref{fig:pie_gpt_pair} and Figure~\ref{fig:pie_gemini_pair} show DeepSeek-V3.1 closely matches human preferences. Against GPT-5.1, AlphaContext wins 62.07\% under human judgments and 60.10\% under DeepSeek-V3.1, with lose rates around 13\%. Against Gemini-3.0-Pro-Preview, AlphaContext wins 73.89\% under human judgments and 67.49\% with DeepSeek, with low lose rates of 2.96\% and 5.91\%. The human–LLM agreement is high (Cohen's $\kappa > 0.8$), supporting the reliability of the LLM judgments for our main evaluations.

\begin{figure}[t]
    \centering
    \begin{subfigure}[b]{0.48\linewidth}
    \centering
    \includegraphics[width=\linewidth, height=\linewidth, keepaspectratio]{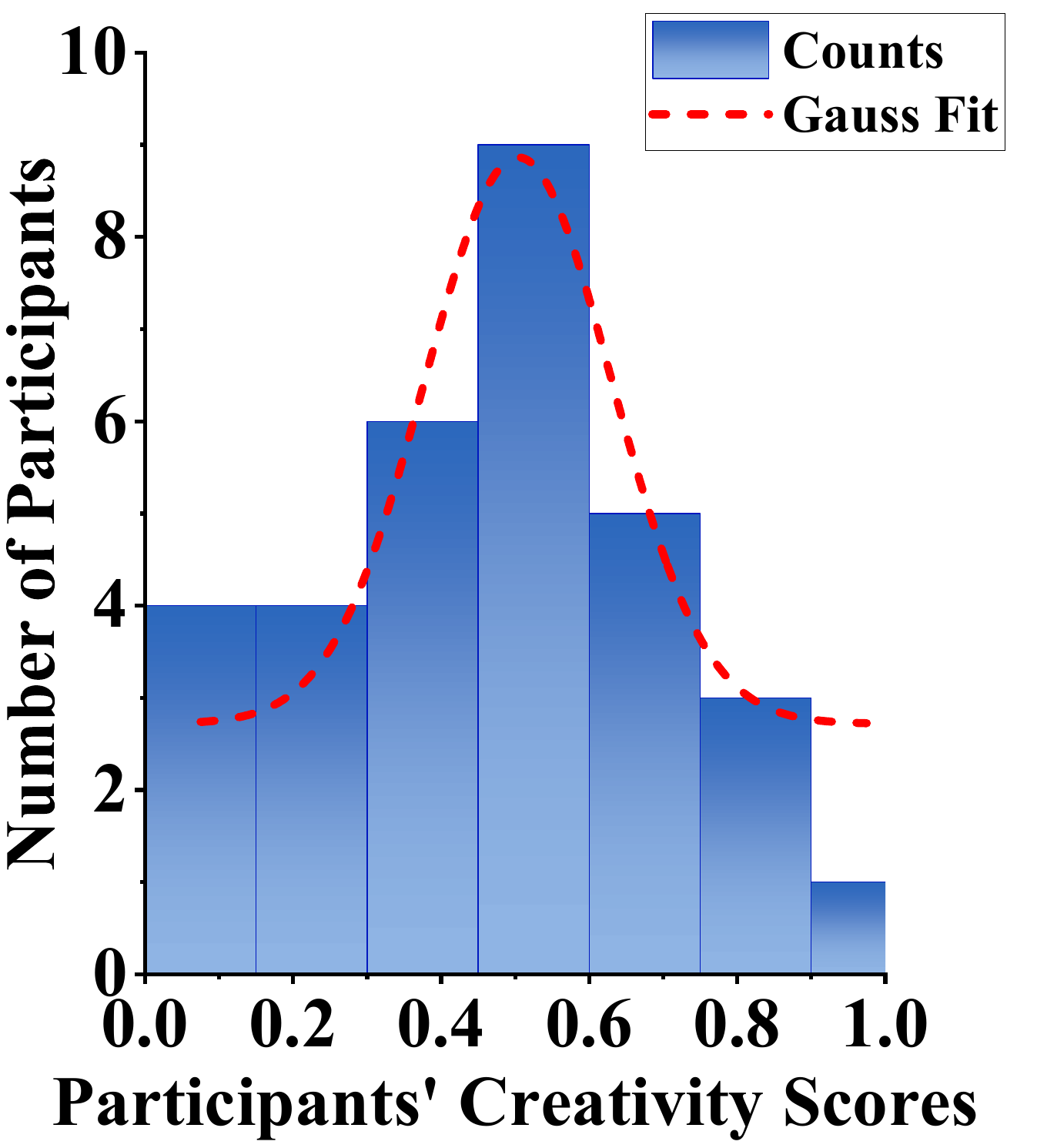}
    \caption{}
    \label{fig:stand_fit}
    \end{subfigure}
    \hfill
    \begin{subfigure}[b]{0.48\linewidth}
    \centering
    \includegraphics[width=\linewidth, height=\linewidth, keepaspectratio]{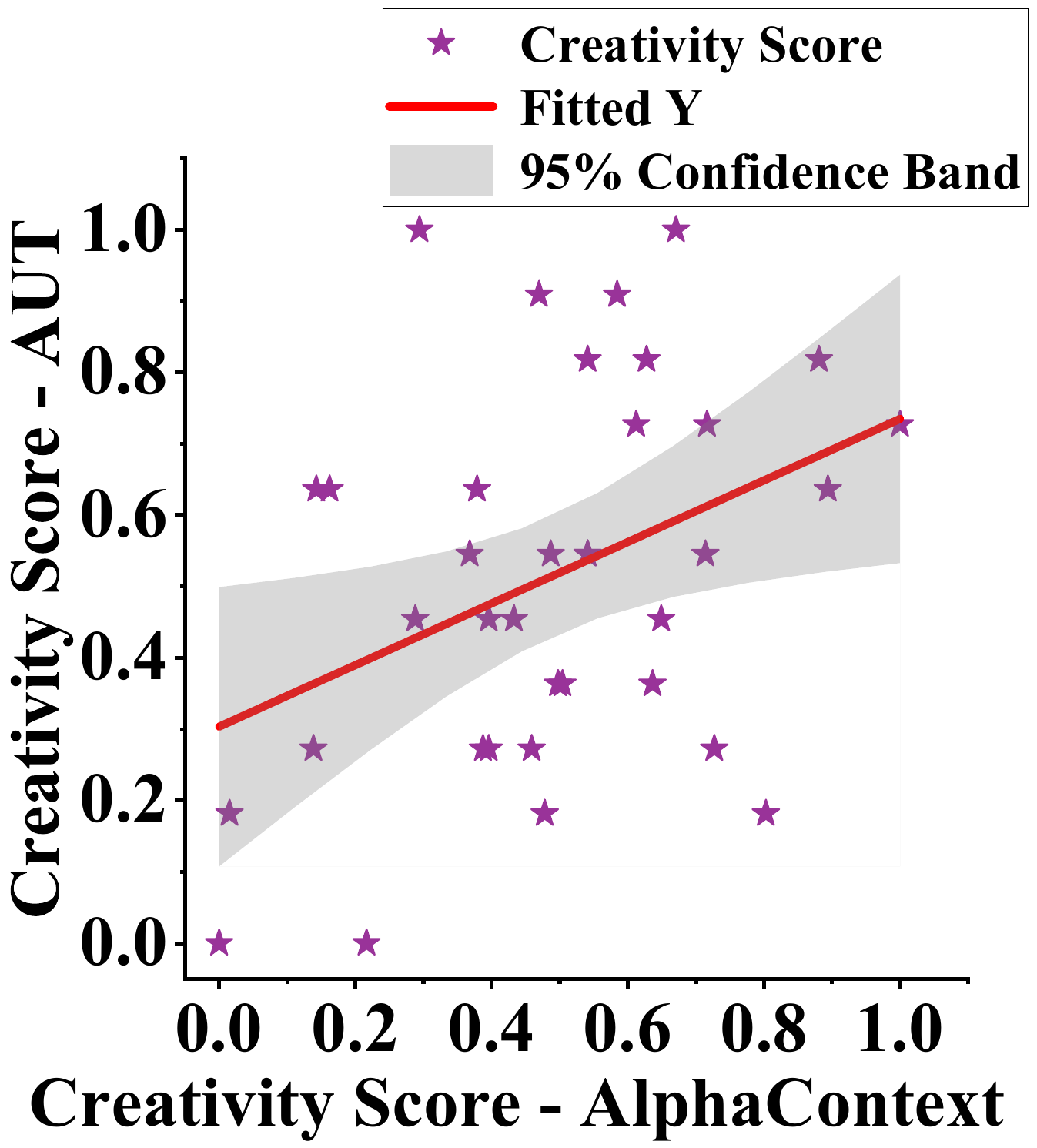}
    \caption{}
    \label{fig:alphacontext_aut_correlation}
    \end{subfigure}
    \caption{Human study results for AlphaContext. (a): distribution of participant creativity scores with a Gaussian fit. (b): Pearson correlation between AlphaContext-based scores and AUT (Alternative Uses Task) scores.} 
    \label{fig:fit and correlation} 
\end{figure}


\noindent\textbf{Real-World Human Study (To Q5).}
We validate the assessment effectiveness of AlphaContext-generated contexts through a real-world study with 36 secondary-school students. As shown in Figure~\ref{fig:stand_fit}, the scores exhibit a symmetric unimodal pattern, and the Gaussian fit closely matches the empirical histogram, suggesting a stable measurement behavior. For criterion validity, we compare AlphaContext-based scores with those of the Alternative Uses Task (AUT), a widely used standardized creativity test. The two scores are collected independently, and the Pearson correlation shows a significant positive association ($r=0.3770$),
as shown in Figure~\ref{fig:alphacontext_aut_correlation}. Notably, according to standard guidelines on psychology and creativity assessment~\cite{gignac2016effect,funder2019evaluating,runco2012divergent,beaty2021automating,beaty2022semantic,benedek2013assessment}, 
a correlation coefficient of $r=0.3770$ is regarded as practically meaningful 
and provides reasonable support for criterion validity. This result indicates that the creativity levels elicited by our generated contexts are consistent with an established benchmark. Overall, the human study provides real-world evidence that AlphaContext can measure student creativity in authentic educational settings. More details can be found in Appendix~\ref{sec:human}.

\begin{figure}[htbp]
    \centering
    \begin{subfigure}[b]{0.48\linewidth}
    \centering
    \includegraphics[width=\linewidth, keepaspectratio]{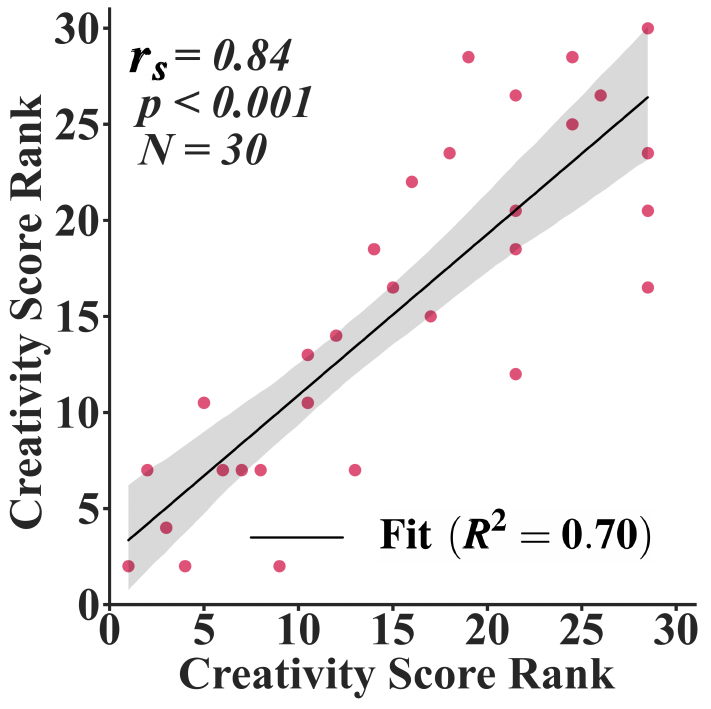}
    \caption{AlphaContext vs. Expert}
    \label{fig:sp1}
\end{subfigure}
\hfill
\begin{subfigure}[b]{0.48\linewidth}
    \centering
    \includegraphics[width=\linewidth, keepaspectratio]{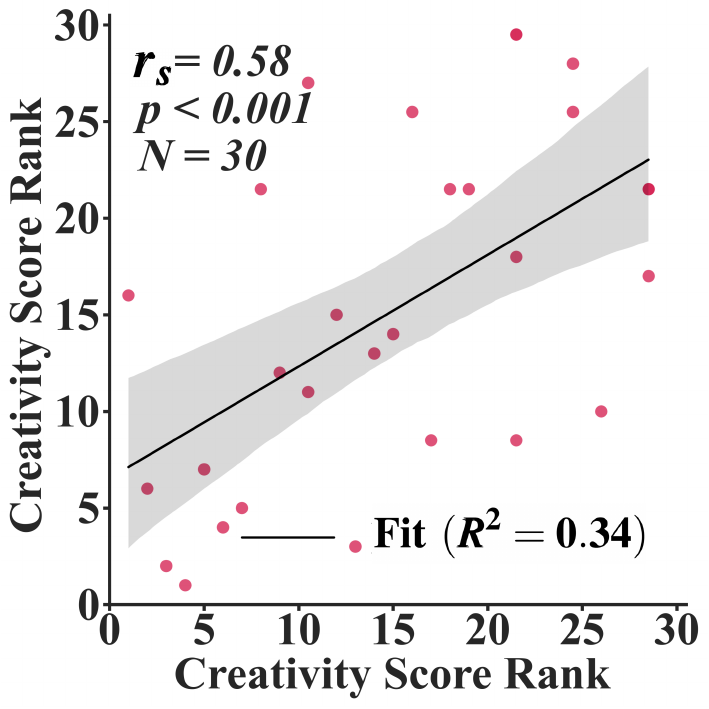}
    \caption{Gemini vs. Expert}
    \label{fig:sp2}
\end{subfigure}
    \caption{Case study on measurement-level alignment with expert assessment. Scatter plots compare expert-induced score ranks (x-axis) with ranks induced by generated contexts (y-axis) for 30 simulated participants.}
    \label{fig:sp_correlation_fitting}
\end{figure}

\noindent\textbf{Case Study (To Q6).}
To evaluate measurement-level alignment with expert assessment, we conduct a controlled case study on the same theme using three contexts: expert-designed, AlphaContext-generated, and Gemini-3-Pro-Preview-generated. We simulate 30 virtual participants with diverse response styles, collect creativity scores, and compare the induced rankings using Spearman correlation and $R^2$ fit. Figure~\ref{fig:sp_correlation_fitting} shows that AlphaContext better matches the expert context ($r_s=0.84$) than Gemini ($r_s=0.58$), indicating closer outcome-level consistency with expert-designed assessments. Details are provided in Appendix~\ref{sec:case}.


\noindent\textbf{Computational Cost Analysis (To Q7).}
We analyze the computational cost of AlphaContext in terms of generation time and token consumption, compared to baseline LLMs and manual expert design. Table~\ref{tab:cost} reports the average time and token usage required to generate one creativity context. Although AlphaContext requires a longer inference time than standard zero-shot prompting, this overhead comes from its full pipeline and is necessary to ensure psychometric validity beyond what simple prompting can reliably provide. In practice, AlphaContext generates one context in about 6 minutes, making it practical for high-quality dataset construction. By contrast, manual expert design typically requires at least one week~\cite{crabbe1989future,barbot2019creativity}. AlphaContext therefore substantially reduces human labor while maintaining quality. In addition, it uses a locally deployed open-source model, avoiding costly closed-source APIs and improving transparency. Overall, the added computational cost is justified by gains in validity, reliability, and automation.

\begin{table}[t]
  \centering
  \small
  \caption{Average generation time and token consumption per context.}
  \label{tab:cost}
  \begin{tabular}{lcc}
    \toprule
    Method & Time (s) & Tokens (k) \\
    \midrule
    GPT-5.1 & 23.50 & 2.36 \\
    Gemini-3.0-Pro-Preview & 46.31 & 3.29 \\
    AlphaContext (Ours) & 226.99 & 12.89 \\
    \bottomrule
  \end{tabular}
\end{table}

\section{Conclusion}
This paper introduces AlphaContext, an evolutionary generator for psychometric assessment contexts that integrates rule-guided outline planning, sentence-level MCTS generation, MAP-Elites quality–diversity optimization, and assessment-guided refinement through virtual participant simulation. Across extensive experiments, AlphaContext consistently outperforms strong LLM baselines and structured frameworks on 7 evaluation dimensions, and shows higher alignment with expert-designed contexts. While it consumes more tokens and requires longer generation time than baseline LLMs, its computational overhead is fully acceptable, and it simultaneously achieves significantly higher assessment validity and generation stability. Human–LLM preference evaluations support reliable automated judging, and a real-world study provides practical validity evidence. Overall, AlphaContext offers a scalable way to produce contexts for creativity assessment while reducing the reliance on scarce expert writing.

AlphaContext is designed as a context generator for human creativity assessment, while also providing standardized contexts for benchmarking LLM creativity. Current experiments focus on future-oriented contexts and a compact expert-curated input set; extending to broader domains, age groups, and languages is an important direction.

\section*{Limitations}
AlphaContext primarily targets generating psychometrically grounded creativity assessment contexts that are usable in real testing settings. However, achieving stable discourse-level coherence and measurement-relevant cue control relies on sentence-level MCTS and MAP-Elites refinement, which require repeated model calls and scoring. As a result, the overall generation cost depends not only on AlphaContext's design but also on the underlying LLM and judge configuration, making efficiency comparisons sensitive to the chosen models and evaluation setup. In addition, psychometrically suitable title–theme inputs are still relatively scarce, which can affect the scale of benchmarking and the extent to which conclusions transfer beyond our current future-oriented setting.

AlphaContext offers strong controllability and assessment alignment, while direct prompting is typically cheaper but less reliable for measurement-oriented constraints. Notably, context generation is not a real-time or online task, and the computational overhead is offset by massive reductions in manual expert effort. In future work, we plan to expand expert-curated inputs and use AlphaContext to produce high-quality training data for fine-tuning of lightweight generators. This will improve efficiency while preserving assessment utility, enabling broader coverage across domains, populations, and deployment settings.

\section*{Ethical Considerations}
This work introduces the CreaTE dataset for creativity-context generation. The title–theme inputs are authored by creativity-psychology experts under a shared specification and contain no personal information. We screen the dataset to remove sensitive or identifiable content, and we prioritize both data quality and ethical compliance during curation. In particular, we apply strict quality-control procedures, including thorough manual review, to ensure broad coverage while proactively addressing potential bias and sensitivity concerns and we curate the dataset in accordance with applicable privacy and research-ethics standards.

Our work also includes a real-world human study to validate the assessment effectiveness of AlphaContext-generated contexts. This study has been reviewed and approved by the Institutional Review Board (IRB) of the affiliated university (IRB Approval No. HR2-0478-2025). Participation was voluntary, and all participants were informed of the study purpose and procedures, with the right to withdraw at any time without penalty. Before the study, we obtained written informed consent from participants and their guardians, and the consent materials specified the study goals, tasks, potential risks, and data use and protection measures. All collected responses were anonymized by removing personal identifiers, stored securely with restricted access, and reported only in aggregate. The study involved minimal risk, as participants completed open-ended creativity tasks similar to typical classroom activities, and we did not request or record sensitive personal information.

\section*{Acknowledgements}
We would like to thank the anonymous reviewers for constructive comments. This work is supported by the National Natural Science Foundation of China (No. 62476091), the General Program in Education of the National Social Science Fund of China (No. BEA230071), and the Key Program in Education of the National Social Science Fund of China (No. ABA220028). 

\bibliography{custom}

@inproceedings{Arena,
  author       = {Tianle Li and
                  Wei{-}Lin Chiang and
                  Evan Frick and
                  Lisa Dunlap and
                  Tianhao Wu and
                  Banghua Zhu and
                  Joseph E. Gonzalez and
                  Ion Stoica},
  title        = {From Crowdsourced Data to High-quality Benchmarks: Arena-Hard and Benchbuilder Pipeline},
  booktitle    = {Forty-second International Conference on Machine Learning},
  year         = {2025},
  address      = {Vancouver, BC, Canada},
}

@article{uncertainty1,
  title={Uncertainty: A Catalyst for Creativity, Learning and Development},
  author={Beghetto, Ronald A and Jaeger, Garrett J},
  journal={Creativity Theory and Action in Education},
  volume = {6},
  year={2022},
}

@article{uncertainty2,
  title={There is No Creativity without Uncertainty: Dubito Ergo Creo},
  author={Beghetto, Ronald A},
  journal={Journal of Creativity},
  volume={31},
  pages={100005},
  year={2021}
}

@article{concreteness,
  title={The Relationship between Contextual Cues in Virtual Environments and Creative Processes},
  author={Guegan, J{\'e}r{\^o}me and Nelson, Julien and Lubart, Todd},
  journal={Cyberpsychology, Behavior, and Social Networking},
  volume={20},
  number={3},
  pages={202--206},
  year={2017},
}

@article{significance1,
  title={Creativity and the Finding and Solving of Real-World Problems},
  author={Okuda, Shawn M and Runco, Mark A and Berger, Dale E},
  journal={Journal of Psychoeducational assessment},
  volume={9},
  number={1},
  pages={45--53},
  year={1991},
}

@article{significance2,
  title={Creative Thinking in the Real World},
  author={Mumford, Michael D and Martin, Robert and Elliott, Samantha and McIntosh, Tristan},
  journal={The nature of human creativity},
  pages={147--65},
  year={2018},
}

@article{creativity1,
  title={The Standard Definition of Creativity},
  author={Runco, Mark A and Jaeger, Garrett J},
  journal={Creativity research journal},
  volume={24},
  number={1},
  pages={92--96},
  year={2012},
}

@article{creativity2,
  title={The Concept of Creativity: Prospects and Paradigms},
  author={Sternberg, Robert J and Lubart, Todd I},
  journal={Handbook of creativity},
  volume={1},
  number={3-15},
  year={1999},
}

@article{creativity-ai,
  title={Rethinking Creativity: Creative Industries, AI and Everyday Creativity},
  author={Lee, Hye-Kyung},
  journal={Media, Culture \& Society},
  volume={44},
  number={3},
  pages={601--612},
  year={2022},
}

@article{creativity-ass,
  title={Creativity Assessment in Psychological Research:(Re)setting the Standards.},
  author={Barbot, Baptiste and Hass, Richard W and Reiter-Palmon, Roni},
  journal={Psychology of Aesthetics, Creativity, and the Arts},
  volume={13},
  number={2},
  pages={233},
  year={2019},
}

@article{sternberg,
  title={Toward a Triarchic Theory of Human Intelligence},
  author={Sternberg, Robert J},
  journal={Behavioral and Brain Sciences},
  volume={7},
  number={2},
  pages={269--287},
  year={1984},
}

@article{future-context,
  title={{A} Time for Creativity: How Future-Oriented Schemas Facilitate Creativity},
  author={Koh, Brandon and Leung, Angela K-y},
  journal={Journal of Experimental Social Psychology},
  volume={84},
  pages={103816},
  year={2019},
}

@article{fpsp,
  title={Creating a Brighter Future: An Update on the Future Problem Solving Program},
  author={Crabbe, Anne Borland},
  journal={Journal for the Education of the Gifted},
  volume={5},
  number={1},
  pages={2--11},
  year={1982},
}

@article{fpsp3,
  title={Interscholastic futuristic creative problem-solving.},
  author={Torrance, E. P. and Bruch, C. B. and Torrance, J. P.},
  journal={The Journal of Creative Behavior},
  volume={10},
  number={2},
  pages={117--125},
  year={1976},
}

@inproceedings{DOC,
  author       = {Kevin Yang and
                  Dan Klein and
                  Nanyun Peng and
                  Yuandong Tian},
  title        = {{DOC:} Improving Long Story Coherence With Detailed Outline Control},
  booktitle    = {Proceedings of the 61st Annual Meeting of the Association for Computational Linguistics},
  pages        = {3378--3465},
  year         = {2023},
  address      = {Toronto, Canada}
}

@article{LiveIdeaBench,
  title={LiveIdeaBench: Evaluating LLMs' Scientific Creativity and Idea Generation with Minimal Context},
  author={Ruan, Kai and Wang, Xuan and Hong, Jixiang and Sun, Hao},
  journal={arXiv e-prints},
  pages={arXiv--2412},
  year={2024}
}

@inproceedings{SiYH25,
  author       = {Chenglei Si and Diyi Yang and Tatsunori Hashimoto},
  title        = {Can LLMs Generate Novel Research Ideas? {A} Large-Scale Human Study with 100+ {NLP} Researchers},
  booktitle    = {Proceedings of the 13th International Conference on Learning Representations},
  year         = {2025},
  address      = {Singapore}
}

@article{Creation-MMbench,
  author       = {Xinyu Fang and
                  Zhijian Chen and
                  Kai Lan and
                  Lixin Ma and
                  Shengyuan Ding and
                  Yingji Liang and
                  Xiangyu Zhao and
                  Farong Wen and
                  Zicheng Zhang and
                  Guofeng Zhang and
                  Haodong Duan and
                  Kai Chen and
                  Dahua Lin},
  title        = {Creation-MMBench: Assessing Context-Aware Creative Intelligence in {MLLM}},
  journal       = {arXiv preprint},
  volume       = {abs/2503.14478},
  year         = {2025},
}

@inproceedings{storyteller,
  author       = {Jiaming Li and
                  Yukun Chen and
                  Ziqiang Liu and
                  Minghuan Tan and
                  Lei Zhang and
                  Yunshui Li and
                  Run Luo and
                  Longze Chen and
                  Jing Luo and
                  Ahmadreza Argha and
                  Hamid Alinejad{-}Rokny and
                  Wei Zhou and
                  Min Yang},
  title        = {{STORYTELLER:} An Enhanced Plot-Planning Framework for Coherent and Cohesive Story Generation},
  booktitle    = {Findings of the Association for Computational Linguistics},
  pages        = {20818--20846},
  year         = {2025},
  address       = {Vienna, Austria}
}

@article{alphaevolve,
  author       = {Alexander Novikov and
                  Ng{\^{a}}n Vu and
                  Marvin Eisenberger and
                  Emilien Dupont and
                  Po{-}Sen Huang and
                  Adam Zsolt Wagner and
                  Sergey Shirobokov and
                  Borislav Kozlovskii and
                  Francisco J. R. Ruiz and
                  Abbas Mehrabian and
                  M. Pawan Kumar and
                  Abigail See and
                  Swarat Chaudhuri and
                  George Holland and
                  Alex Davies and
                  Sebastian Nowozin and
                  Pushmeet Kohli and
                  Matej Balog},
  title        = {AlphaEvolve: {A} coding agent for scientific and algorithmic discovery},
  journal      = {arXiv preprint},
  volume       = {arXiv:2506.13131},
  year         = {2025},
}

@inproceedings{item_generator,
  author       = {Antonio Laverghetta Jr. and
                  Simone Luchini and
                  Averie Linnell and
                  Roni Reiter{-}Palmon and
                  Roger E. Beaty},
  title        = {The Creative Psychometric Item Generator: a Framework for Item Generation and Validation Using Large Language Models},
  booktitle    = {Proceedings of the 3rd Workshop on Artificial Intelligence and Creativity co-located with 27th European Conference on Artificial Intelligence},
  pages        = {59--73},
  year         = {2024},
  address      ={Santiago de Compostela, Spain}
}

@inproceedings{structure_story,
  author       = {Angela Fan and
                  Mike Lewis and
                  Yann N. Dauphin},
  title        = {Strategies for Structuring Story Generation},
  booktitle    = {Proceedings of the 57th Conference of the Association for Computational Linguistics},
  pages        = {2650--2660},
  year         = {2019},
  address      = {Florence, Italy}
}

@inproceedings{longwriter,
  author       = {Yushi Bai and
                  Jiajie Zhang and
                  Xin Lv and
                  Linzhi Zheng and
                  Siqi Zhu and
                  Lei Hou and
                  Yuxiao Dong and
                  Jie Tang and
                  Juanzi Li},
  title        = {LongWriter: Unleashing 10,000+ Word Generation from Long Context LLMs},
  booktitle    = {Proceedings of the 13th International Conference on Learning Representations},
  year         = {2025},
  address      = {Singapore}
}

@inproceedings{CRITICS,
  author       = {Minwook Bae and Hyounghun Kim},
  title        = {Collective Critics for Creative Story Generation},
  booktitle    = {Proceedings of the 2024 Conference on Empirical Methods in Natural Language Processing},
  pages        = {18784--18819},
  year         = {2024},
  address      = {Miami, FL, {USA}}
}

@inproceedings{SS-GEN,
  author       = {Yi Feng and
                  Mingyang Song and
                  Jiaqi Wang and
                  Zhuang Chen and
                  Guanqun Bi and
                  Minlie Huang and
                  Liping Jing and
                  Jian Yu},
  title        = {{SS-GEN:} {A} Social Story Generation Framework with Large Language Models},
  booktitle    = {AAAI-25, Sponsored by the Association for the Advancement of Artificial Intelligence},
  pages        = {1300--1308},
  year         = {2025},
  address      = {Philadelphia, PA, {USA}}
}

@article{Neural_representations,
  author={Wang, Xueyang and Liu, Wei and Zhuang, Kaixiang and Liu, Cheng and Zhang, Jingyi and Fan, Li and Chen, Qunlin and Qiu, Jiang},
  title={Neural representations of noncentral events during narrative encoding predict subsequent story ending originality},
  journal={Science Advances},
  volume= {11},
  number= {17},
  pages = {eadu5251},
  year={2025},
}

@article{LLM_thinking,
  author       = {Kuang{-}Huei Lee and
                  Ian Fischer and
                  Yueh{-}Hua Wu and
                  Dave Marwood and
                  Shumeet Baluja and
                  Dale Schuurmans and
                  Xinyun Chen},
  title        = {Evolving Deeper {LLM} Thinking},
  journal      = {arXiv preprint },
  volume       = {arXiv:2501.09891},
  year         = {2025}
}

@inproceedings{drama,
  title        = {Towards Enhanced Immersion and Agency for LLM-based Interactive Drama},
  author       = {Hongqiu Wu and
                  Weiqi Wu and
                  Tianyang Xu and
                  Jiameng Zhang and
                  Hai Zhao},
  booktitle    = {Proceedings of the 63rd Annual Meeting of the Association for Computational Linguistics},
  pages        = {11166--11182},
  year         = {2025},
  address      = {Vienna, Austria}
}

@inproceedings{CollabLLM,
  author       = {Shirley Wu and
                  Michel Galley and
                  Baolin Peng and
                  Hao Cheng and
                  Gavin Li and
                  Yao Dou and
                  Weixin Cai and
                  James Zou and
                  Jure Leskovec and
                  Jianfeng Gao},
  title        = {CollabLLM: From Passive Responders to Active Collaborators},
  booktitle    = {Proceedings of the 42nd International Conference on Machine Learning},
  year         = {2025},
  address      = {Vancouver, Canada}
}

@inproceedings{aidanbench,
   title={AidanBench: Stress-Testing Language Model Creativity on Open-Ended Questions}, 
   author={Aidan Mclaughlin and James Campbell and Anuja Uppuluri and Yiming Yang},
   booktitle={NeurIPS 2024 Workshop on Language Gamification},
   year={2024}
}

@article{automated,
  title={Automated Scoring of Creative Problem Solving with Large Language Models: A Comparison of Originality and Quality ratings.},
  author={Luchini and Simone A and Maliakkal and Nadine T and DiStefano and Paul V and Laverghetta Jr and Antonio and Patterson and John D and Beaty and Roger E and Reiter-Palmon, Roni},
  journal={Psychology of Aesthetics, Creativity, and the Arts},
  year={2025},
}

@article{scope,
  title={Milieus of Creativity: The Role of Places, Environments, and Spatial contexts},
  author={Meusburger, Peter},
  journal={Milieus of creativity: An interdisciplinary approach to spatiality of creativity},
  volume={2},
  pages={97--153},
  year={2009},
}

@article{knowledge,
  title={Idea Density and the Creativity of Written Works},
  author={Runco, Mark A and Turkman, Burak and Acar, Selcuk and Nural, Mustafa V},
  journal={Journal of Genius and Eminence},
  volume={2},
  number={1},
  pages={26--31},
  year={2017}
}

@article{viewpoint,
  title={Intellectual and Viewpoint Diversity: Importance, Scope and Bounds},
  author={VanderWeele, Tyler J},
  journal={Education Sciences},
  volume={15},
  number={12},
  pages={1592},
  year={2025},
}

@article{crabbe1989future,
  title={The Future Problem Solving Program},
  author={Crabbe, A.B.},
  journal={Educational Leadership},
  volume={7},
  number={1},
  pages={27--29},
  year={1989}
}

@article{barbot2019creativity,
  title={Creativity assessment in psychological research:(Re) setting the standards},
  author={Barbot, B. and Hass, R.W. and Reiter-Palmon, R.},
  journal={Psychology of Aesthetics, Creativity, and the Arts},
  volume={13},
  number={2},
  pages={233},
  year={2019}
}

@article{gignac2016effect,
  title={Effect size guidelines for individual differences researchers},
  author={Gignac, G.E. and Szodorai, E.T.},
  journal={Personality and Individual Differences},
  volume={102},
  pages={74--78},
  year={2016}
}

@article{funder2019evaluating,
  title={Evaluating effect size in psychological research: Sense and nonsense},
  author={Funder, D.C. and Ozer, D.J.},
  journal={Advances in Methods and Practices in Psychological Science},
  volume={2},
  number={2},
  pages={156--168},
  year={2019}
}

@article{runco2012divergent,
  title={Divergent thinking as an indicator of creative potential},
  author={Runco, M.A. and Acar, S.},
  journal={Creativity Research Journal},
  volume={24},
  number={1},
  pages={66--75},
  year={2012}
}

@article{beaty2021automating,
  title={Automating creativity assessment with SemDis: An open platform for computing semantic distance},
  author={Beaty, R.E. and Johnson, D.R.},
  journal={Behavior Research Methods},
  volume={53},
  number={2},
  pages={757--780},
  year={2021}
}

@article{beaty2022semantic,
  title={Semantic distance and the alternate uses task: Recommendations for reliable automated assessment of originality},
  author={Beaty, R.E. and Johnson, D.R. and Zeitlen, D.C. and Forthmann, B.},
  journal={Creativity Research Journal},
  volume={34},
  number={3},
  pages={245--260},
  year={2022}
}

@article{benedek2013assessment,
  title={Assessment of divergent thinking by means of the subjective top-scoring method: Effects of the number of top-ideas and time-on-task on reliability and validity},
  author={Benedek, M. and Mühlmann, C. and Jauk, E. and Neubauer, A.C.},
  journal={Psychology of Aesthetics, Creativity, and the Arts},
  volume={7},
  number={4},
  pages={341},
  year={2013}
}

@article{amabile1983social,
  title={The social psychology of creativity: A componential conceptualization},
  author={Amabile, T.M.},
  journal={Journal of Personality and Social Psychology},
  volume={45},
  number={2},
  pages={357},
  year={1983}
}

@book{amabile2018creativity,
  title={Creativity in context: Update to the social psychology of creativity},
  author={Amabile, T.M.},
  publisher={Routledge},
  year={2018}
}

@inproceedings{SAPD,
  author       = {Yixuan Wang and
                  Jiale Feng and
                  Yue Huang and
                  Xuruo Pan and
                  Zhongjing Huang and
                  Zhi Liu and
                  Hong Qian},
  title        = {{A} Style-Aware Polytomous Diagnostic Model for Individual Traits},
  booktitle    = {Proceedings of the 28th European Conference on Artificial Intelligence},
  pages        = {2698--2705},
  year         = {2025},
  address      = {Bologna, Italy}
}

@article{guo,
  title={Twenty-first century creativity: An investigation of how the partnership for 21st century instructional framework reflects the principles of creativity},
  author={Guo, Jiajun and Woulfin, Sarah},
  journal={Roeper Review},
  volume={38},
  number={3},
  pages={153--161},
  year={2016},
}

@article{zhong1,
  title={Random tree model of meaningful memory},
  author={Zhong, Weishun and Can, Tankut and Georgiou, Antonis and Shnayderman, Ilya and Katkov, Mikhail and Tsodyks, Misha},
  journal={Physical Review Letters},
  volume={134},
  number={23},
  pages={237402},
  year={2025},
}

@article{zhong2,
  author       = {Weishun Zhong and
                  Doron Sivan and
                  Tankut Can and
                  Mikhail Katkov and
                  Misha Tsodyks},
  title        = {Semantic Chunking and the Entropy of Natural Language},
  journal      = {CoRR},
  volume       = {abs/2602.13194},
  year         = {2026},
}

\appendix

\section*{Appendix}
\label{sec:appendix}
\section{CreaTE Dataset}
\label{sec:dataset}
AlphaContext takes a title and a theme as input, so evaluation requires inputs that are explicitly designed for creativity assessment. Existing story-generation datasets are ill-suited for this purpose: they target narrative completeness and stylistic richness, but do not ensure that the topic focus, implicit cues, and task intent satisfy psychometric requirements. Using such datasets would confound the evaluation, as poor performance could arise from mismatched inputs rather than the generation method.

\begin{figure}[htbp]
    \centering
    \includegraphics[width=1.0\linewidth]{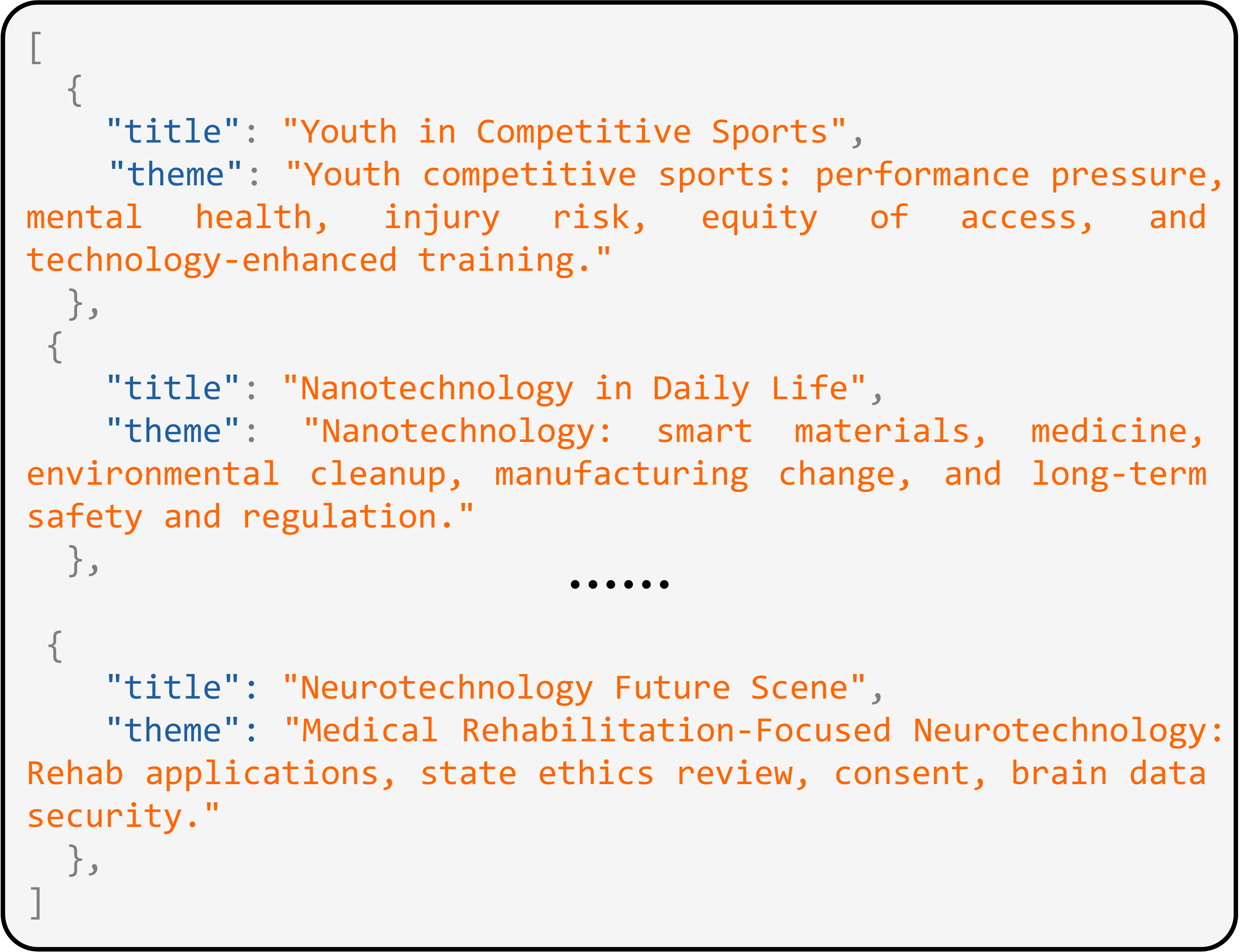} 
    \caption{Example input format of CreaTE. Each entry includes a title and a theme, designed to support psychometrically grounded creativity context generation.}
    \label{fig:create_example}
\end{figure}

To address this gap, we curate CreaTE, a compact yet high-quality input set for creativity-context generation. CreaTE contains 203 title–theme pairs spanning diverse domains, balancing evaluation cost with broad coverage. Each entry is authored by three creativity-psychology experts under a shared specification: the title provides a concrete anchor, while the theme delineates the intended problem space and key creative tensions to be elicited, as shown in Figure~\ref{fig:create_example}. A subset of expert-designed titles and themes is adapted from publicly available FPSP topic resources\footnote{\url{https://fpspi.org/topics/}}. We refine the dataset through cross-checking and iterative revision, and include an entry only after expert consensus on assessment relevance, clarity, and correctness, with edits applied to remove ambiguity or unintended cues. Finally, we conduct compliance screening to ensure that no sensitive or identifiable information is present.

\section{Preference Evaluation}
\label{sec:gemini}

We provide additional details on the human evaluation protocol and inter-annotator agreement analysis for the preference study involving Gemini-3.0-Pro-Preview.

For human judgments, we recruited three evaluators and adopted a standardized rating protocol with a pre-study calibration session. The human evaluation checklist is strictly aligned with our metric definitions (Coherence, Relevance, Engagement, Significance, Concreteness, Uncertainty) to ensure consistent interpretation. Inter-rater agreement among the three evaluators meets the required consistency standard, and we report the mean of their judgments as the final human preference result.

Consistent with observations in the main text, DeepSeek's judgments closely track human preferences, with high human–LLM agreement (Cohen's $\kappa>0.8$), further validating the reliability of automated evaluation in our experiments.


\section{Cue Coverage and Diversity in a Case Comparison}
\begin{figure}[htbp]
    \centering
    \includegraphics[width=1.0\linewidth]{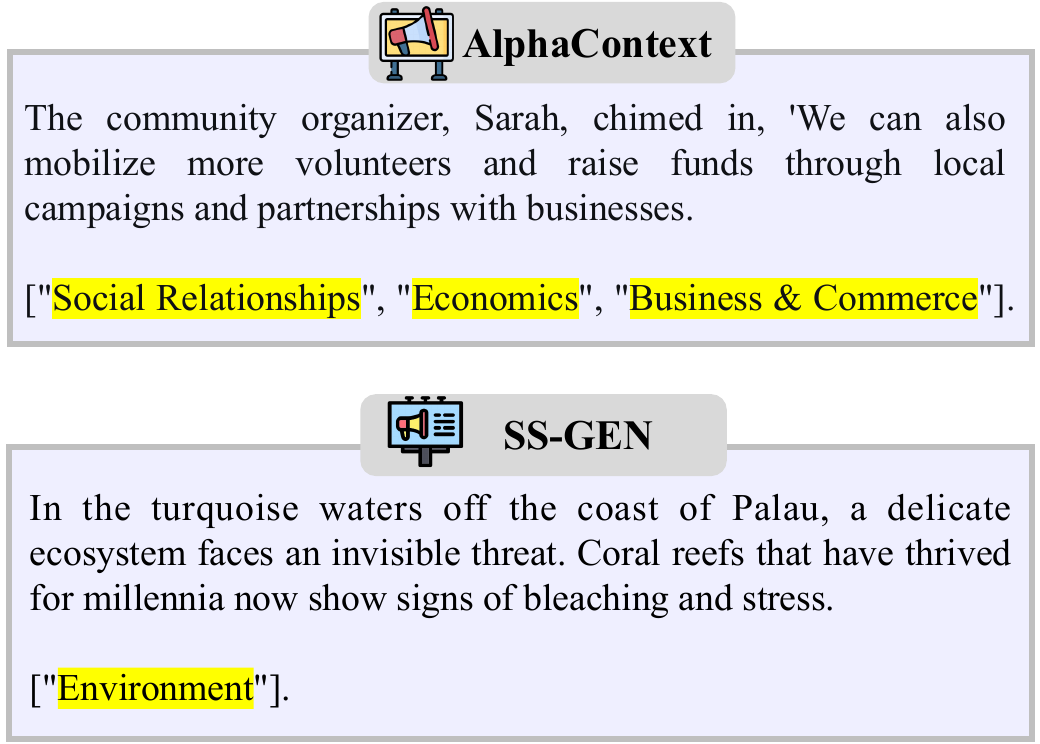} 
    \caption{Case comparison under the same input theme: AlphaContext embeds a broader and more diverse set of assessment cues in a coherent fragment, whereas SS-GEN concentrates on a single dominant cue, resulting in narrower cue coverage.}
    \label{fig:cue}
\end{figure}

Under the same input theme, AlphaContext produces a context fragment with noticeably broader assessment-cue coverage than SS-GEN. Figure~\ref{fig:cue} shows that AlphaContext can surface multiple challenge dimensions within a single coherent narrative move, so the scenario invites reasoning from several angles rather than focusing on only one. In contrast, SS-GEN typically centers the context around a single dominant cue, which yields a narrower cue footprint and fewer directions for subsequent idea exploration. This qualitative comparison aligns with our design goal: by planning cue placement at the outline level and enforcing outline-grounded generation, AlphaContext increases the diversity of assessment-relevant cues while keeping them implicitly integrated into the story, better supporting creativity assessment that aims to elicit open-ended and multi-perspective thinking.

\section{Detailed Analysis of Human Study}
\label{sec:human}
To complement the aggregate Pearson correlation reported in the \textit{Real-World Human Study}, we further examine how specific creativity dimensions align between AlphaContext and a standardized benchmark using Spearman rank correlation. We used Pearson correlation in the main text because the aggregated total scores empirically exhibit an approximately unimodal, near-Gaussian distribution, making Pearson an appropriate summary of linear association. In contrast, dimension-level creativity scores can be more heterogeneous in both cognitive mechanisms and distributional shapes. We therefore adopt Spearman correlation to provide a rank-based and more robust analysis of whether the two assessments preserve consistent relative ordering at a finer granularity.

Before the formal study, all participants completed the assessment under a standardized administration procedure to ensure comparability across individuals. The instructions, task materials, and time constraints were fixed and delivered in a consistent format. Participants were guided to complete the tasks independently and were discouraged from discussion or external assistance during the session. To protect privacy, we removed personally identifiable information from all collected records, used anonymized participant identifiers for subsequent analysis, and stored the data in an access-restricted manner. Only de-identified responses and scores were used for reporting and correlation analysis.

All human scores used in this analysis were produced under a standardized expert rating protocol. Specifically, three psychology experts served as raters. A domain specialist first developed a detailed scoring rubric and dimension definitions, and then conducted fine-grained training for the raters before annotation. The training session explained each dimension with concrete guidance and calibration examples, ensuring a shared understanding of the scoring criteria and common failure cases. During scoring, the three raters evaluated responses independently. We then computed inter-rater agreement to verify consistency, and the agreement met the required standard. Finally, we report the mean score across the three raters as the human score for each dimension.

\begin{figure}[ht]
    \centering
    \centering
    \includegraphics[width=\linewidth, keepaspectratio]{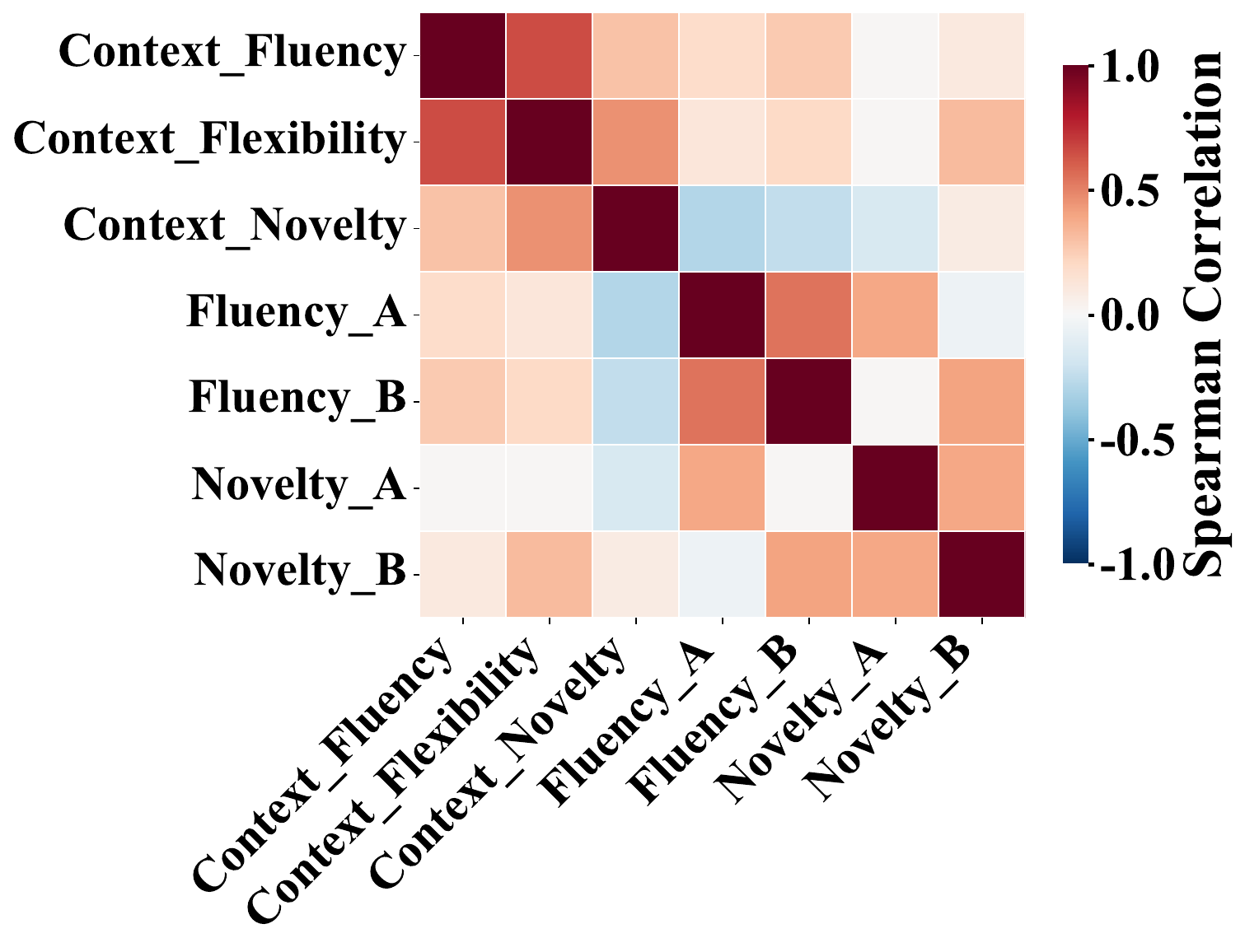}
    \caption{Spearman rank-correlation heatmap between AlphaContext metrics and AUT metrics.}
    \label{fig:aut_heatmap}
\end{figure}

Figure~\ref{fig:aut_heatmap} shows the Spearman correlation matrix between AlphaContext dimension scores and Alternative Uses Task (AUT) scores. The AUT metrics are reported for two AUT prompts, denoted by the suffixes \texttt{\_A} and \texttt{\_B}. Overall, the heatmap suggests a meaningful but nuanced alignment. In particular, \texttt{Context\_Fluency} exhibits a consistent positive association with AUT fluency, supporting validity on ideational productivity. Meanwhile, correlations involving \texttt{Context\_Novelty} are weaker and less directly matched to the standard AUT novelty. This pattern is expected and informative: AUT measures unconstrained divergent thinking, whereas AlphaContext evaluates creativity elicited under explicit semantic and contextual constraints. The results indicate that our method aligns with established benchmarks on general fluency-related signals, while capturing a context-dependent aspect of creativity that may not be fully reflected by standard AUT scoring.

\section{Extended Measurement-level Alignment Study Across Baselines}
\label{sec:case}
To complement the main case study (To Q6), we extend the same measurement-level alignment analysis to all baselines. For each method, we generate a context under the same theme as the expert-designed reference, simulate responses from the same set of 30 virtual participants with diverse response styles, and compute creativity scores under each context. We then compare the outcome-level ranking induced by each method with the expert-induced ranking.

\begin{figure}[htbp]
    \centering
    \centering
    \includegraphics[width=\linewidth, keepaspectratio]{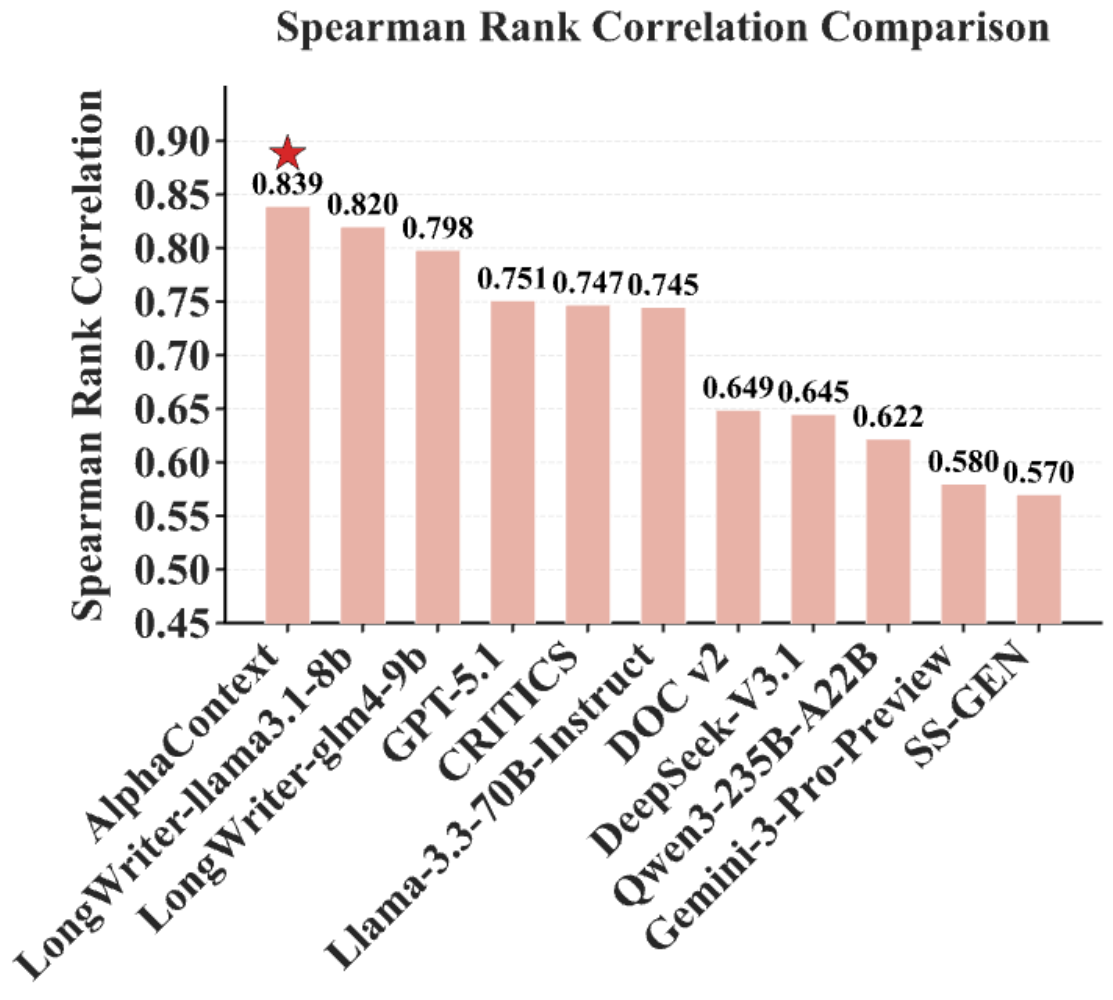}
    \caption{Extended Measurement-level Alignment Study Across Baselines.}
    \label{fig:apx}
\end{figure}

Figure~\ref{fig:apx} summarizes the alignment results across baselines. We report Spearman rank correlation to quantify ranking consistency with the expert reference, and we also include the corresponding $R^2$ from a linear fit as a complementary indicator of overall association. Overall, AlphaContext achieves the strongest agreement with the expert reference, suggesting that it best preserves the relative creativity differences elicited by expert-designed measurement at the outcome level. Several long-form or structurally guided baselines show moderate alignment, whereas general-purpose one-shot generators exhibit noticeably weaker agreement. This baseline-wide comparison provides a broader view of measurement consistency and further supports AlphaContext's advantage beyond single-baseline comparisons.

\section{Analysis of MCG Evaluator}
\label{sec:mcts}
This section analyzes the evaluator used in the MCTS-based Context Generator (MCG). MCG formulates long-form context generation as a sentence-level tree search under a planned outline. Each node represents a partial context, and candidate continuations are explored through MCTS. To decide which branches to expand and keep, the MCG relies on an evaluator that scores each newly generated sentence fragment with respect to creativity-assessment context quality.

The evaluator considers three criteria. Cue alignment $S_{\mathrm{sc}}$ measures how well the generated sentence matches the outline and embeds assessment-relevant cues. It focuses on faithful adherence to the outline hints, including required themes, keywords, constraints, and intended challenge categories, while encouraging implicit planting of meaningful problem cues such as trade-offs, constraints, stakeholder tensions, and second-order effects, instead of explicitly listing challenges. Imagery vividness $S_{\mathrm{im}}$ measures how strongly the text evokes a vivid mental image and immersion through concrete situational and sensory details, including visual impressions, sounds, smells, physical sensations, and emotions, following creativity measurement literature~\cite{Neural_representations}. Discourse coherence $S_{\mathrm{co}}$ evaluates whether the fragment reads smoothly within the evolving context, with stable entities, natural transitions, and clear causal and temporal continuity.

\begin{figure}[htbp]
    \centering
    \includegraphics[width=0.80\linewidth]{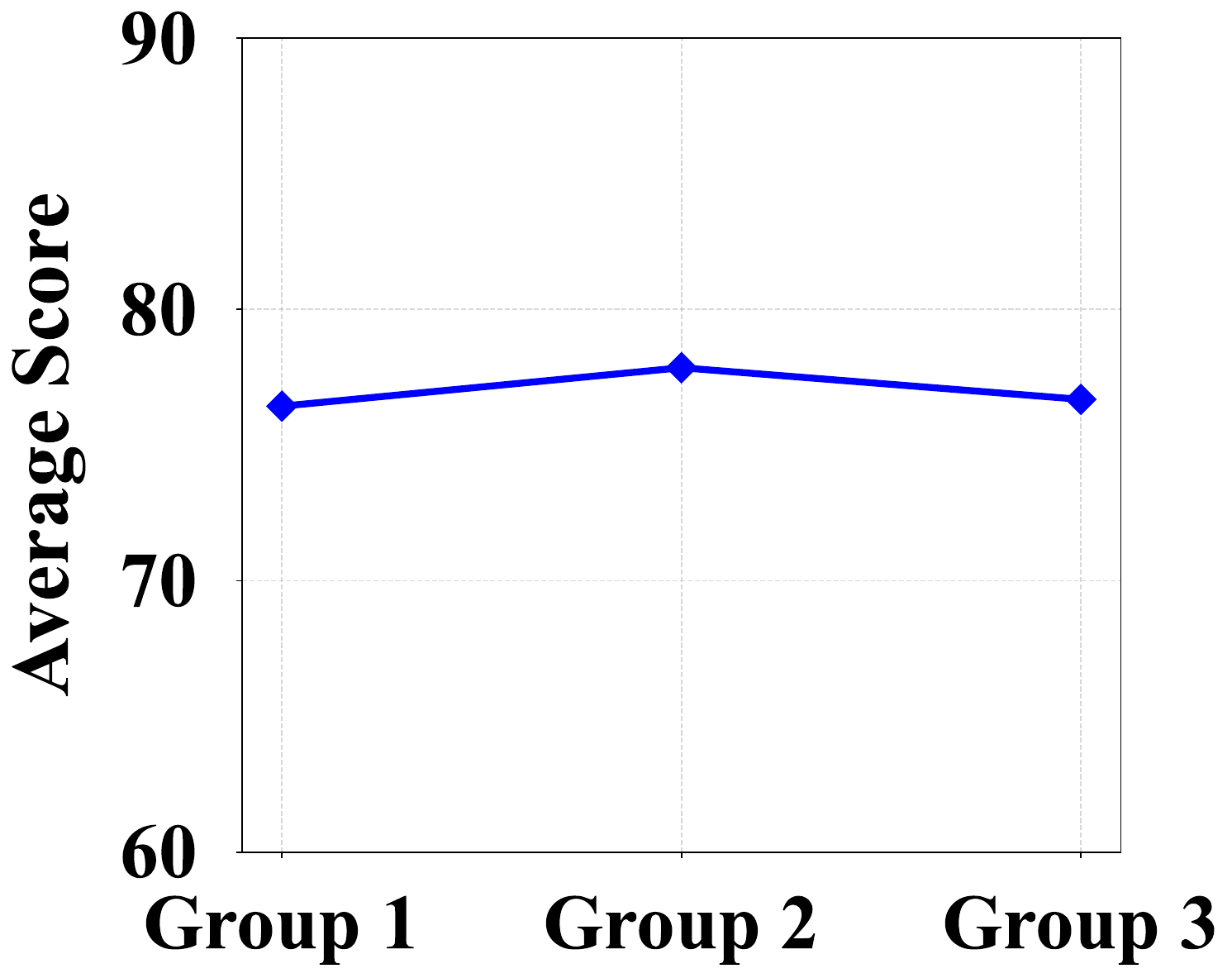} 
    \caption{Comparison of the overall average scores across the three weight groups. }
    \label{fig:average_score}
\end{figure}

\begin{figure}[t]
    \centering
    \begin{subfigure}[b]{0.49\linewidth}
        \centering
        \includegraphics[width=\linewidth]{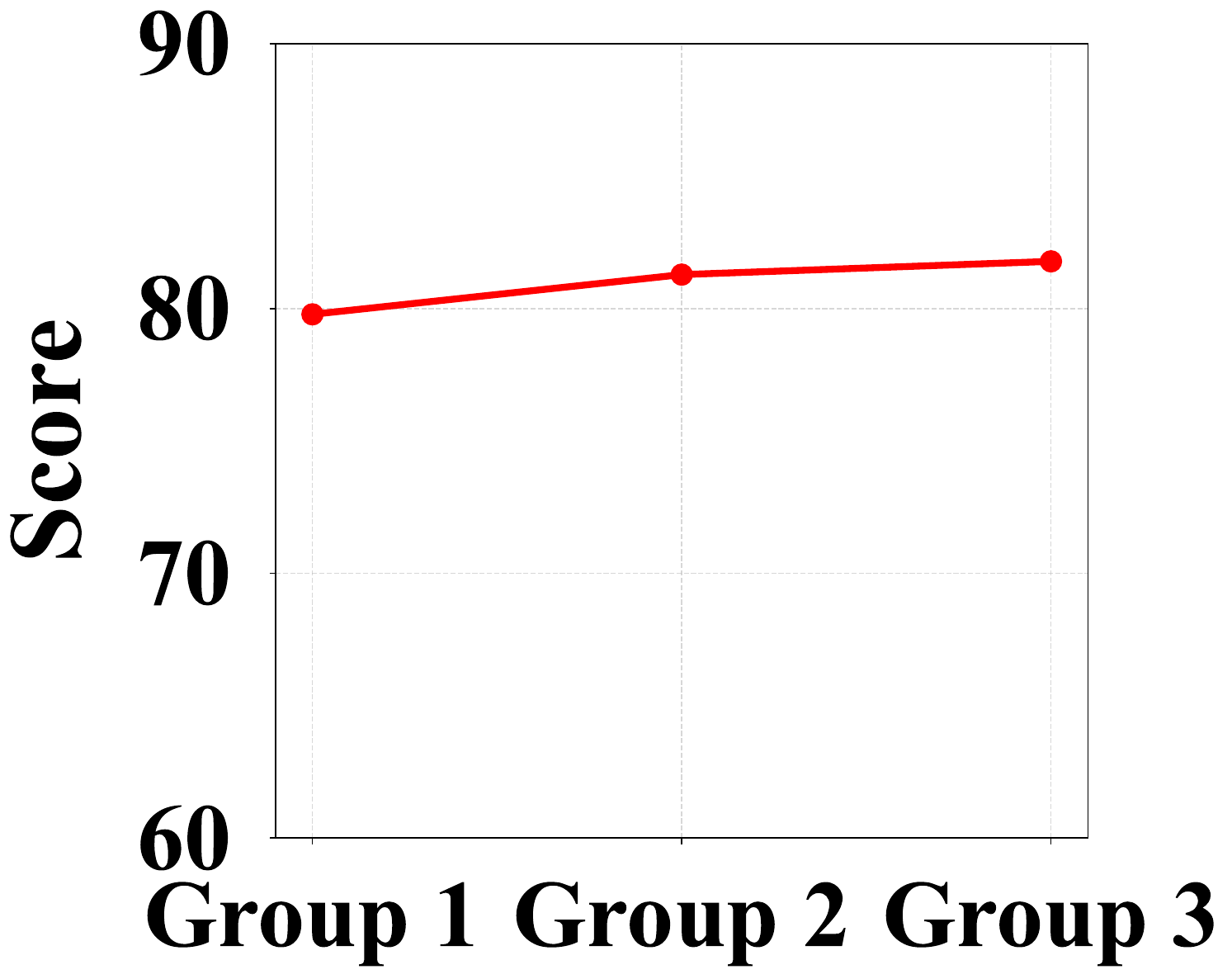}
        \caption{Coherence}
        \label{fig:sub_coherence}
    \end{subfigure}
    \hfill
    \begin{subfigure}[b]{0.49\linewidth}
        \centering
        \includegraphics[width=\linewidth]{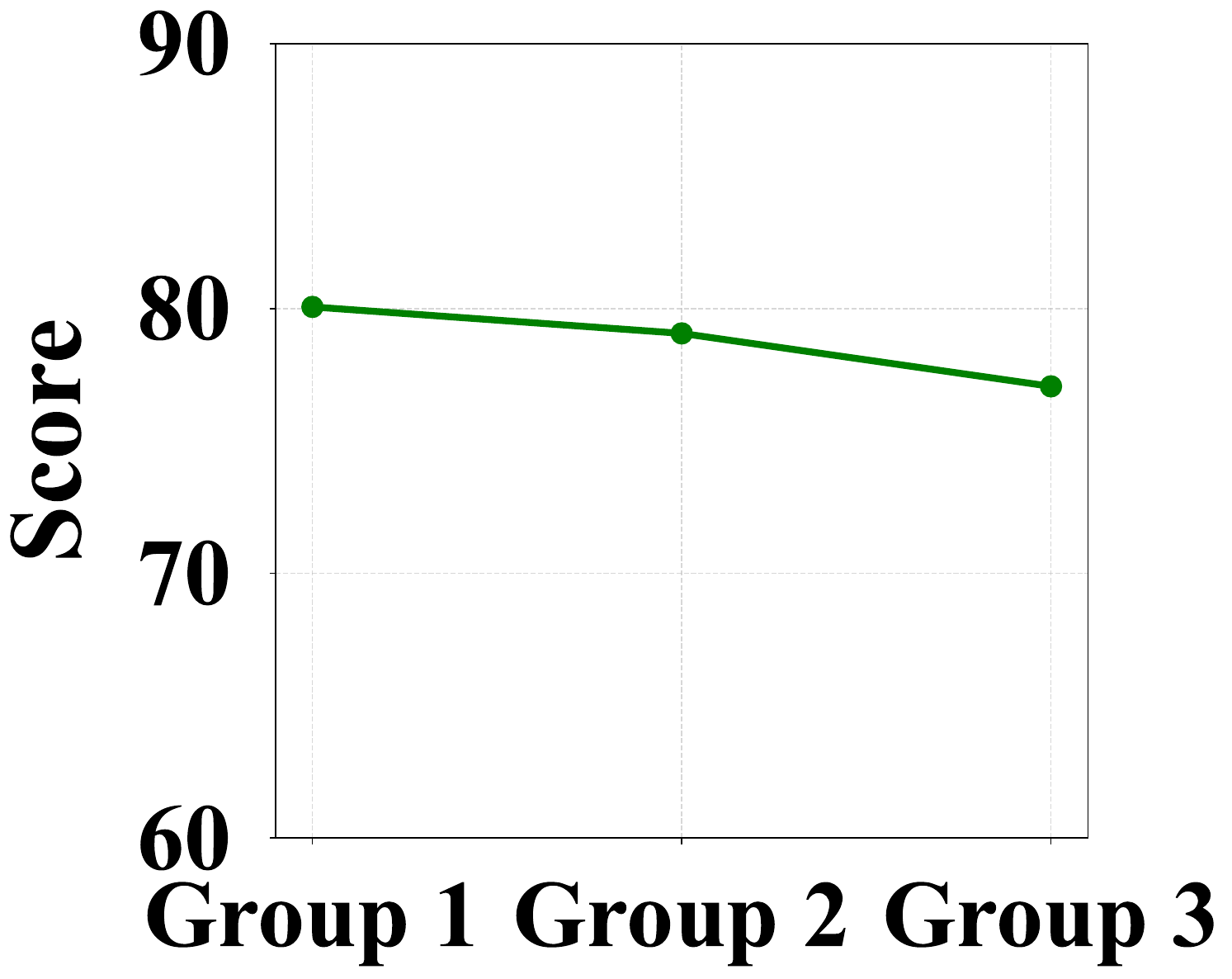}
        \caption{Relevance}
        \label{fig:sub_relevance}
    \end{subfigure}

    \begin{subfigure}[b]{0.49\linewidth}
        \centering
        \includegraphics[width=\linewidth]{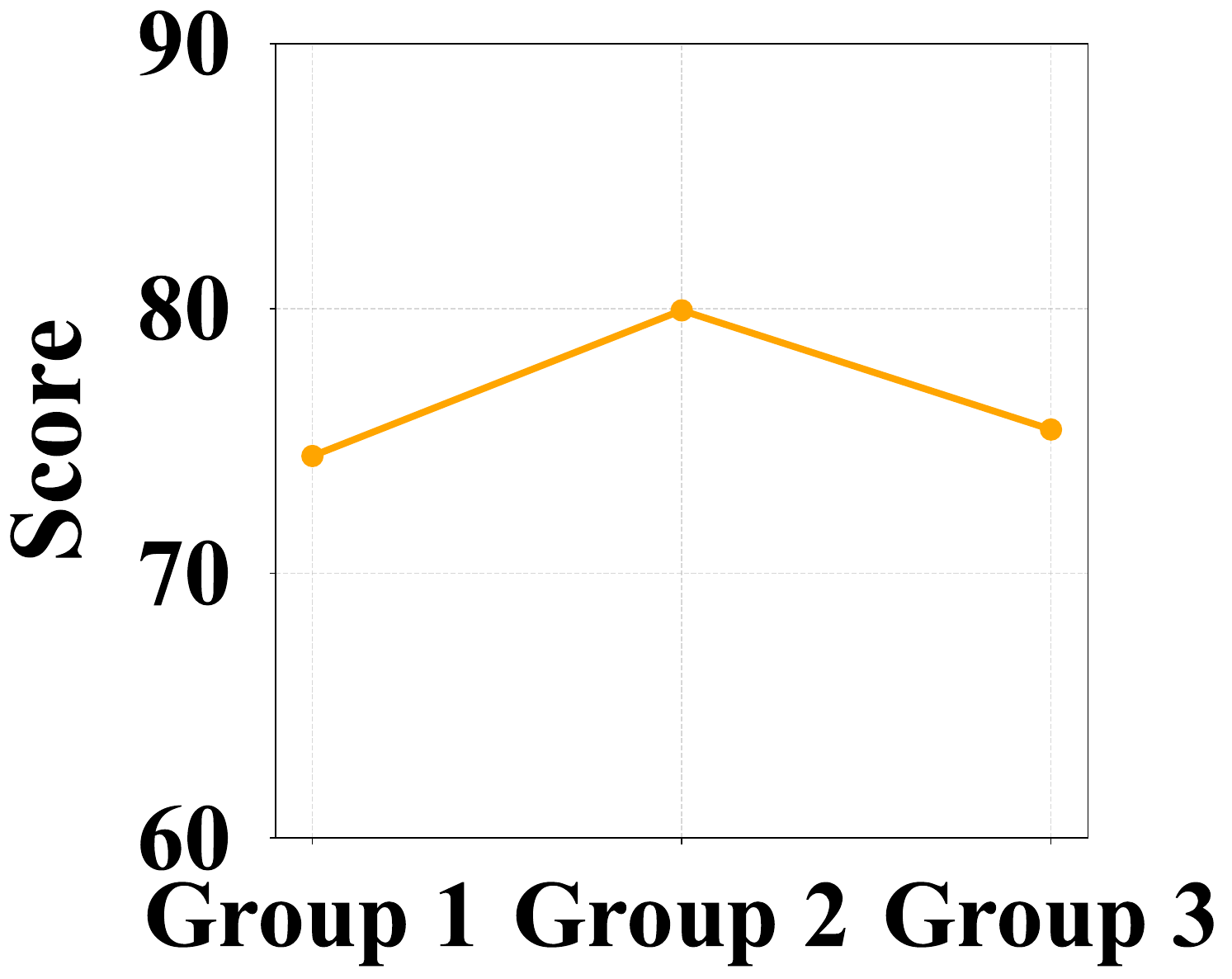}
        \caption{Engagement}
        \label{fig:sub_engagement}
    \end{subfigure}
    \hfill
    \begin{subfigure}[b]{0.49\linewidth}
        \centering
        \includegraphics[width=\linewidth]{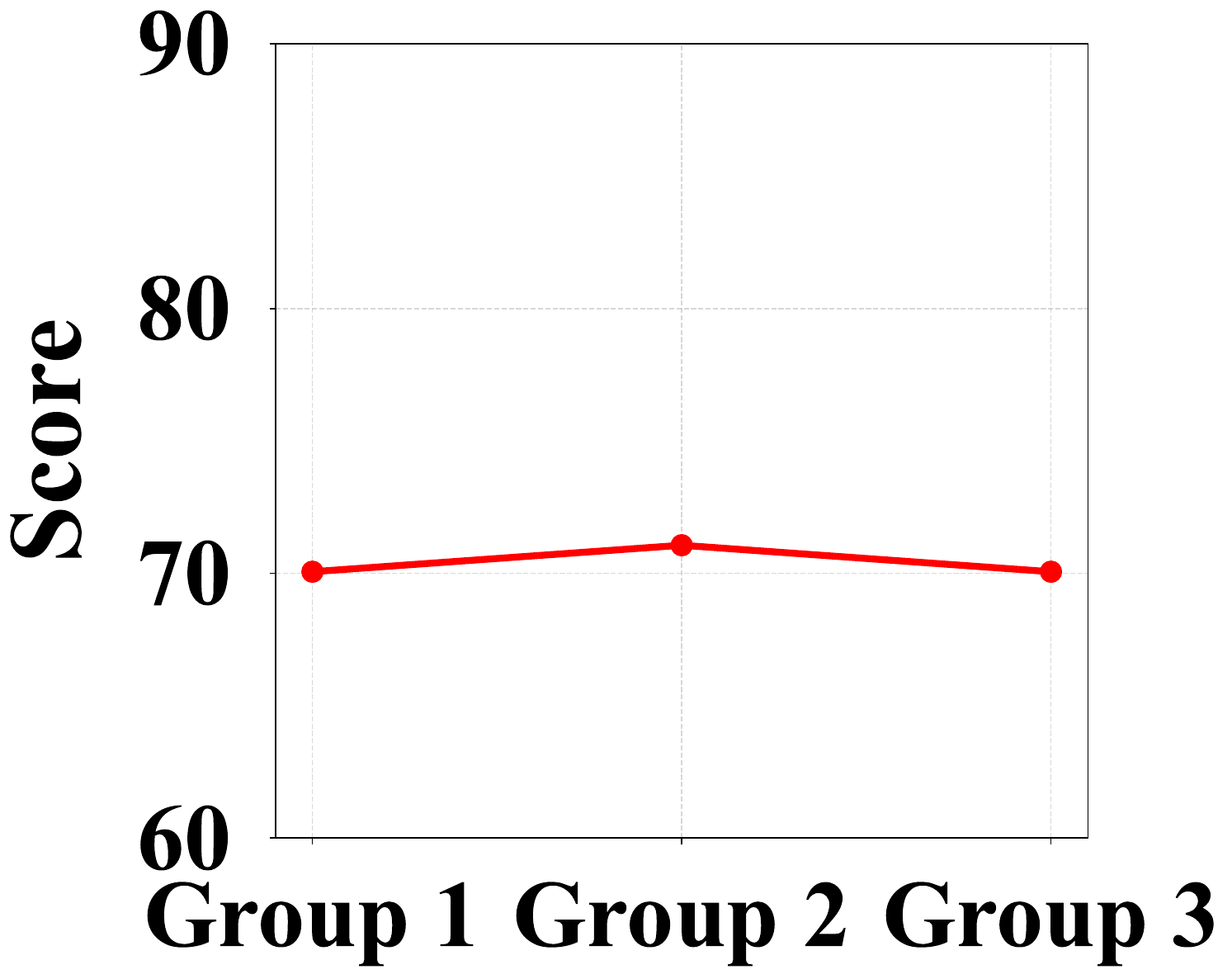}
        \caption{Significance}
        \label{fig:sub_significance}
    \end{subfigure}

    \begin{subfigure}[b]{0.49\linewidth}
        \centering
        \includegraphics[width=\linewidth]{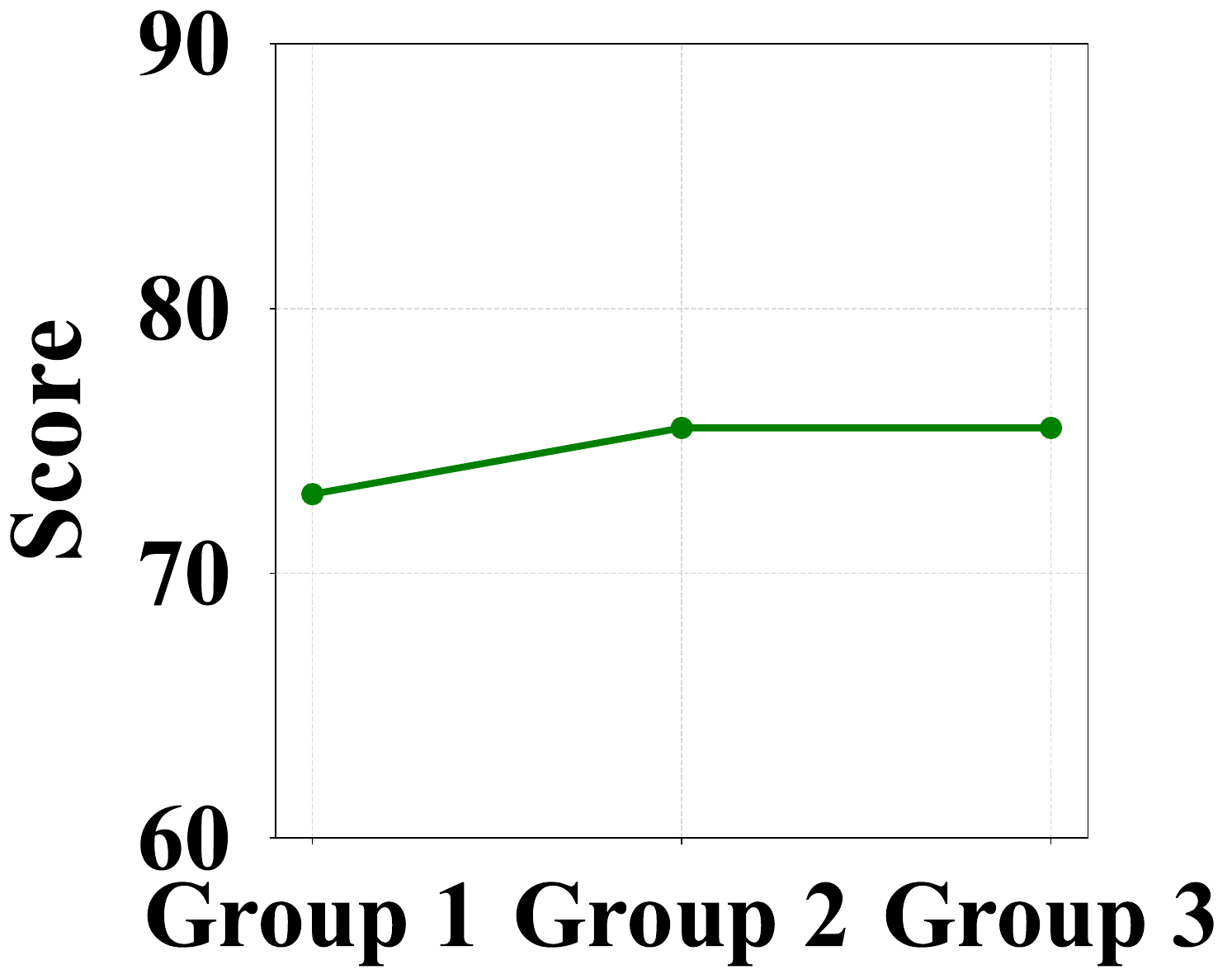}
        \caption{Concreteness}
        \label{fig:sub_concreteness}
    \end{subfigure}
    \hfill
    \begin{subfigure}[b]{0.49\linewidth}
        \centering
        \includegraphics[width=\linewidth]{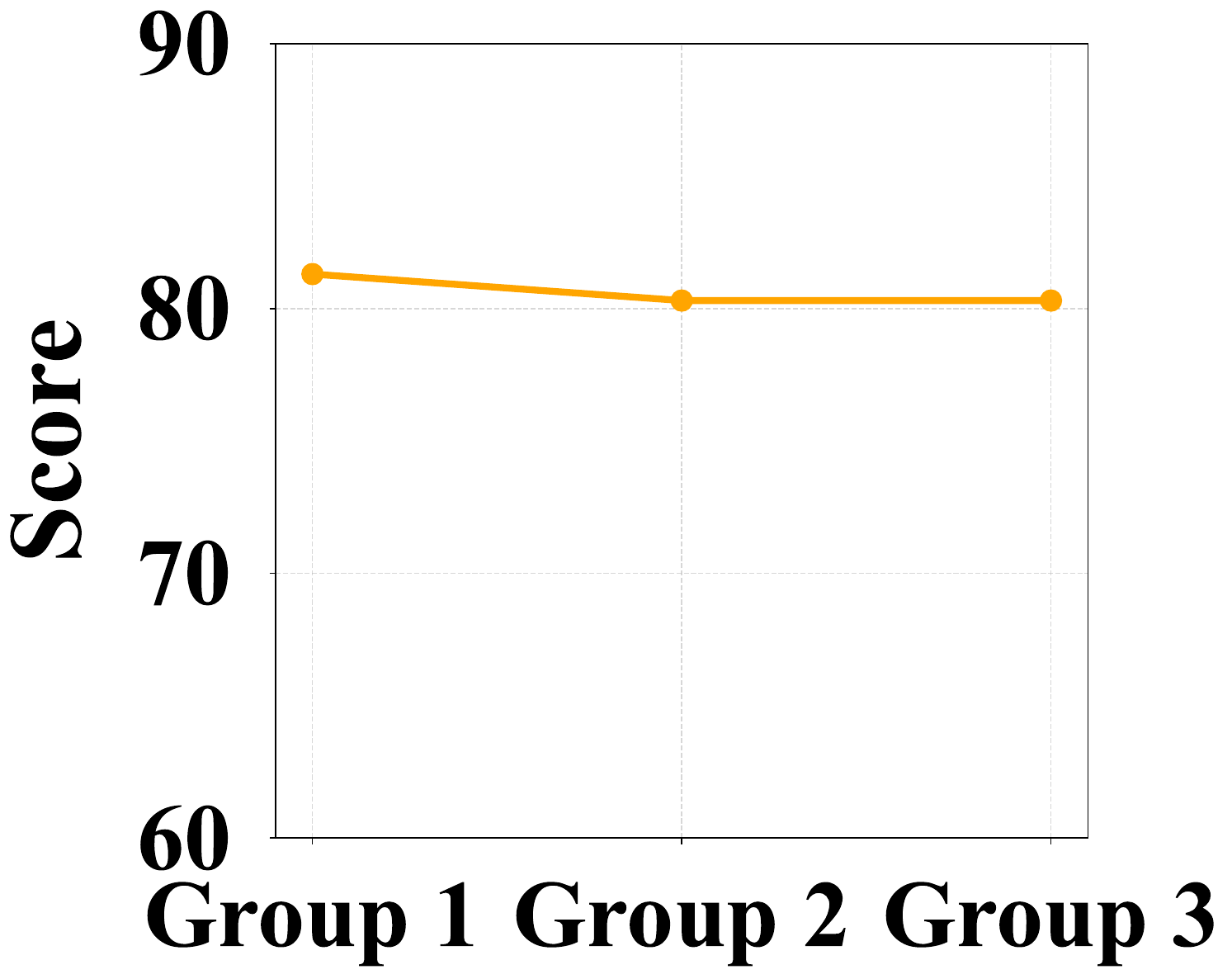}
        \caption{Uncertainty}
        \label{fig:sub_uncertainty}
    \end{subfigure}

    \caption{Effect of different MCG evaluator coefficient settings on the six subjective metrics. Subfigures (a)--(f) report the mean scores for Coherence, Relevance, Engagement, Significance, Concreteness, and Uncertainty, respectively.}
    \label{fig:metrics_stability}
\end{figure}

For evaluation reliability, each criterion is rated on a 5-point Likert scale and then normalized to $[0,1]$, where $L$ is the Likert score. The three normalized scores are aggregated with coefficients $\omega_1,\omega_2,\omega_3$, consistent with the main text, to form the scalar evaluation value used by MCG during the search. We examine three coefficient groups to assess sensitivity: Group 1 uses $\omega_1=\omega_2=\omega_3=0.33$; Group 2 uses $\omega_1=0.4,\ \omega_2=0.3,\ \omega_3=0.3$; Group 3 uses $\omega_1=0.5,\ \omega_2=0.25,\ \omega_3=0.25$. As shown in Figure~\ref{fig:average_score} and Figure~\ref{fig:metrics_stability}, the six subjective metrics exhibit only mild fluctuations across groups, indicating stable evaluation behavior under reasonable coefficient changes. Group 2 yields the highest overall average score, suggesting that moderately emphasizing cue alignment best balances outline-grounded cue placement with vivid writing and coherent discourse. We therefore adopt $\omega_1=0.4,\ \omega_2=0.3,\ \omega_3=0.3$ as the default setting for all subsequent experiments.

\section{Details of the HyperTree Outline Planner}
\label{sec:apphtp}
This section provides additional details for the HyperTree Outline Planner (HTP), complementing the description in Section~\ref{sec:htp}. We focus on how HTP operationalizes outline design as hypertree search, where directed hyperedges enable hierarchical divide-and-conquer over discourse structure and assessment-cue placement. Beyond the high-level phases (HT-Select, HT-Expand, HT-Construct, and HT-Decide), we clarify the rule system that governs admissible expansions and explain how LLM-based decisions are integrated to control search scale and semantic validity.

Section~\ref{subsec:rules} specifies the construction rules of HTP. It includes a static hierarchical skeleton that defines the admissible outline topology, and a set of dynamic, LLM-driven selection rules that prune candidate hyperchains and choose the next divisible leaf node to expand. Section~\ref{subsec:case_study} then presents a fully instantiated outline for the title \textbf{\textit{AI Partner}}, illustrating how these rules concretely materialize into a structured outline suitable for creativity assessment context generation.

\subsection{HyperTree Construction Rules}
\label{subsec:rules}
HTP follows a hybrid rule mechanism that combines a predefined context-free grammar with semantic-aware dynamic selection. The context-free grammar provides a static skeleton that constrains the outline to valid hierarchical forms, while dynamic selection rules use an LLM to evaluate candidate hyperchains under a budget and to select the most promising divisible node for expansion. This hybrid design separates structural admissibility from semantic quality control, allowing HTP to maintain global validity while remaining flexible to the input title and theme. The formal definitions of the static skeleton and the dynamic selection nodes are provided in Listing~\ref{lst:rules}.

\begin{figure*}[t]
\begin{lstlisting}[caption={Formal definitions of static rules and dynamic LLM-driven selection rules in the HyperTree Outline Planner.}, label={lst:rules}, captionpos=b]
# ==================== Part 1: Fixed Hierarchical Skeleton (Static Rules) ==================

1. [Plan] -> [Anchor][Scene Setting][Characters & Interaction][Conflict & Challenge][Open Task] # The plan can be divided into five core narrative aspects.
2. [Anchor] -> [Future Horizon][Place][Scale][Challenge Seeds 1] 
    # Establishes the fundamental time, space, and scope coordinates.
3. [Scene Setting] -> [Scenario Frame][Constraint Hints][Challenge Seeds 2] 
    # Defines context and constraints.
4. [Characters & Interaction] -> [Interaction Goal][Dispute Focus][Problem Slot][Challenge Seeds 3] 
    # Constructs interpersonal dynamics.
5. [Conflict & Challenge] -> [Challenge Seeds 4][Creativity Triggers] 
    # Introduces complicating factors.
6. [Open Task] -> [Challenge Identification][Solution Exploration] 
    # Defines the student's objective.

# ==================== Part 2: Dynamic Selection Nodes (LLM-Driven) =======================
    # Logic A: When this node is selected for expansion, the LLM selects one option from the predefined candidate pool based on theme relevance.
9.  [Future Horizon] -> {NearFuture (5-15y) | MidFuture | FarFuture | Speculative}
10. [Scale] -> {Community | National | International | Space}
11. [Scenario Frame] -> {Everyday Life | Urban Infrastructure | Virtual-Reality Fusion | ...} 

12. [Interaction Goal] -> {Co-creation Workshop | Negotiation | Emergency Response | ...}
13. [Dispute Focus] -> {Value Conflict | Resource Conflict | Trust Conflict | ...}
14. [Creativity Triggers] -> {Uncertainty | Contradiction | Resource Constraints | ...}
    
    # Logic B: When this node is selected for expansion, the LLM selects multiple options from the predefined candidate pool based on theme relevance.
15. [Challenge Seeds 1] -> {{Select 2-3 seeds from Pool}} 
16. [Challenge Seeds 2] -> {{Select 2-3 seeds from Pool}} 
17. [Challenge Seeds 3] -> {{Select 3-4 seeds from Pool}} 
18. [Challenge Seeds 4] -> {{Select 4-5 seeds from Pool}}

19. [Topic Phrase] -> {{LLM-generated phrase (6-8 words)}} # Summarizes the core conflict based on Title/Theme.

20. [Constraint Hints] -> {{Select 2-3 from: Policy, Budget, Time Limit, Safety, etc.}} # Limits the solution space.
\end{lstlisting}
\end{figure*}

\section{Details of the Evolutionary Context Optimizer}
This section provides additional implementation details of the Evolutionary Context Optimizer, with a focus on how we operationalize the style-oriented behavior space and how the MAP-Elites archive is constructed and updated. Initialized with the seed contexts produced by the MCTS-based Context Generator, this module searches in natural language space to jointly expand stylistic coverage and improve within-niche quality, yielding a diverse and assessment-ready context archive $\mathcal{A}$.

\subsection{Style-Oriented Behavior Space}
To characterize stylistic variations that are highly relevant to future-problem contexts for creativity assessment, we map each candidate context $C$ into a task-specific behavior space $B$ via the descriptor $b(C)$. Following the formulation in Eq.~\ref{eq8}, $b(C)$ consists of three interpretable dimensions. The first dimension, $\phi_1(C)$, captures \emph{proximity scope}, measuring how the context is framed from personal daily-life settings to broader public and societal issues~\cite{scope}. The second dimension, $\phi_2(C)$, captures \emph{knowledge density}, reflecting how strongly the narrative is grounded in objective evidence such as data cues, mechanistic explanations, and causal constraints~\cite{knowledge}. The third dimension, $\phi_3(C)$, captures \emph{viewpoint diversity}, indicating the breadth of stakeholders represented in the context and whether multi-perspective considerations jointly shape the problem space~\cite{viewpoint}. Together, these three axes form a controllable and interpretable coordinate system to organize stylistic diversity in creativity assessment contexts.

\subsection{Descriptor Evaluation, Mutation, and Archive Update}
In our implementation, the three behavior dimensions are computed by a fixed-template LLM-based rater, which takes the context text as input and returns normalized scores for $\phi_1(C)$, $\phi_2(C)$, and $\phi_3(C)$ in a structured JSON format. We then discretize the continuous behavior space uniformly to construct a 3D grid archive, where each cell corresponds to a behavioral niche and stores the current elite context.

To enable controllable style shifts in natural language, we implement mutation as a conditional LLM editing process. Each mutation step specifies target values along the three behavior axes and guides the model to revise the context through insertion, deletion, and replacement, so that the candidate moves toward the desired niche while maintaining narrative readability and cue traceability for assessment use. After mutation, each candidate is evaluated in two aspects. Behavior feature evaluation determines its niche assignment via $b(C)$. Quality evaluation is produced by an LLM-based scorer over coherence, relevance, and engagement, and we compute fitness as the uniform average of these three normalized scores to avoid introducing extra hyperparameters. The archive is updated with niche-wise elite replacement: a candidate is inserted if the niche is empty, and otherwise it replaces the current elite only when it achieves higher fitness.

\subsection{Iteration Budget}
The archive is constructed iteratively. At each iteration, we sample elites from the current archive, generate niche-targeted mutants, evaluate their behavior descriptors and quality scores, and update the archive accordingly. The process terminates when the preset iteration budget is reached, resulting in a context archive that covers diverse stylistic regions while maintaining stable quality within each niche.

\subsection{Case Study: Instantiated HyperTree Outline for AI Partner}
\label{subsec:case_study}

Listing~\ref{lst:case_study} presents a fully expanded HyperTree outline produced by HTP for the theme \textit{Human--AI companionship and autonomy}. The outline is constructed by iteratively applying the expansion rules in Section~\ref{subsec:rules}, resulting in a valid hypertree whose branches correspond to alternative hyperchains and whose leaves specify the finest-grained discourse units. Each leaf node is instantiated into a concrete narrative element that can be directly consumed by the downstream context generator, including assessment-relevant cue carriers such as \textit{Trust Conflict} and grounded setting components such as \textit{Everyday Life}. This example illustrates how the rule system yields an explicit, structured outline that supports controlled cue placement while maintaining flexibility in narrative realization.

\begin{figure*}[t]
\begin{lstlisting}[caption={Instantiated HyperTree outline for the \textit{AI Partner} theme, generated by the HyperTree Outline Planner.}, label={lst:case_study}, captionpos=b]
Title: AI Partner
Theme: Human-AI companionship and autonomy: ethics, emotional reliance, privacy, and governance of pervasive personal AI assistants.

Outline Structure:
                  [Plan]
                    [Anchor]
                      [Future Horizon]
                        [NearFuture (5-15 years)]
                      [Place]
                        [City Or Region]
                        [Specific Facility]
                      [Scale]
                        [Community]
                      [Challenge Seeds 1]
                        [Technology]
                        [Ethics & Morality]
                        [Psychological Health]
                    [Scene Setting]
                      [Scenario Frame]
                        [Everyday Life]
                      [Constraint Hints]
                        [Policy]
                        [Budget]
                        [Time Limit]
                      [Challenge Seeds 2]
                        [Technology]
                        [Ethics & Morality]
                    [Characters & Interaction]
                      [Interaction Goal]
                        [Negotiation Meeting]
                      [Dispute Focus]
                        [Trust Conflict]
                      [Problem Slot]
                        AI autonomy in human-AI emotional companionship
                      [Challenge Seeds 3]
                        [Technology]
                        [Ethics & Morality]
                        [Psychological Health]
                    [Conflict & Challenge]
                      [Challenge Seeds 4]
                        [Technology]
                        [Ethics & Morality]
                        [Psychological Health]
                        [Social Relationships]
                        [Law & Justice]
                      [Creativity Triggers]
                        [Uncertainty Cue]
                        [Contradiction Cue]
                    [Open Task]
                      [Challenge Identification]
                        [Prompt student to identify multiple challenges in the scenario]
                      [Solution Exploration]
                        [Prompt student to think of possible response strategies]
\end{lstlisting}
\end{figure*}


\section{Prompt Templates}
\label{sec:prompt}
In this section, we provide the prompt templates used in this study, including the chat template and the evaluation template.

\subsection{Chat Template}
\label{sec:chat_prompt}
\begin{figure*}[t]
  \centering
    \includegraphics[width=\linewidth]{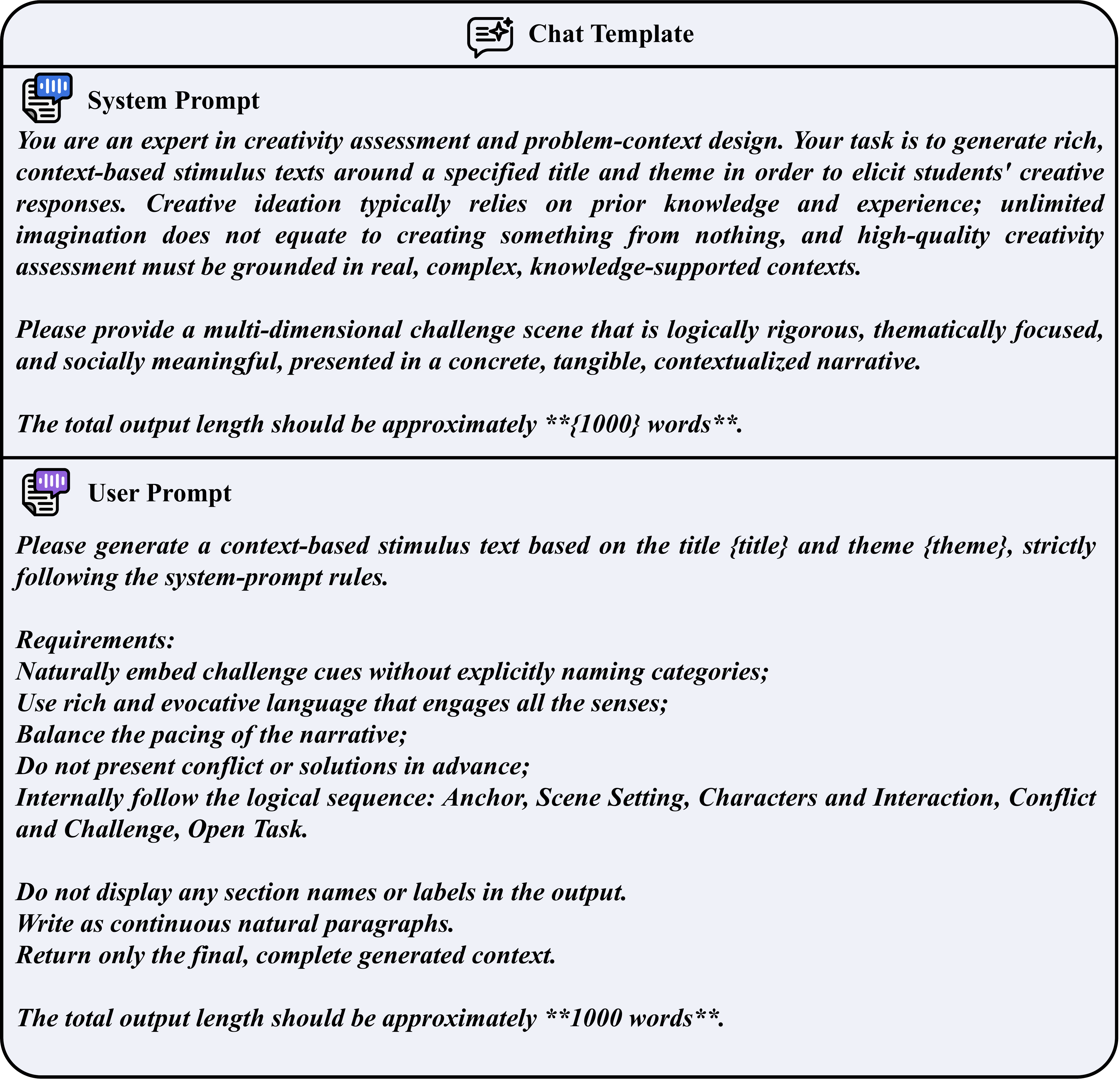}
    \caption{Unified chat prompt template used by baseline LLMs for creativity context generation.}
    \label{fig:prompt}
\end{figure*}

For DeepSeek-V3.1, Qwen3-235B-A22B, Llama3.3-70B-Instruct, GPT-5.1, Gemini-3.0-Pro-Preview, LongWriter-Llama3.1-8B, and LongWriter-GLM4-9b, we adopt a unified chat template for context generation, as shown in Figure~\ref{fig:prompt}. In this template, the system prompt specifies the model's role and the psychometric constraints required for creativity assessment contexts, whereas the user prompt instantiates the input title and theme and enforces requirements on output formatting and discourse progression.

\subsection{Evaluation Template}
\label{sec:eva_prompt}
\begin{figure*}[t]
  \centering
    \includegraphics[width=\linewidth]{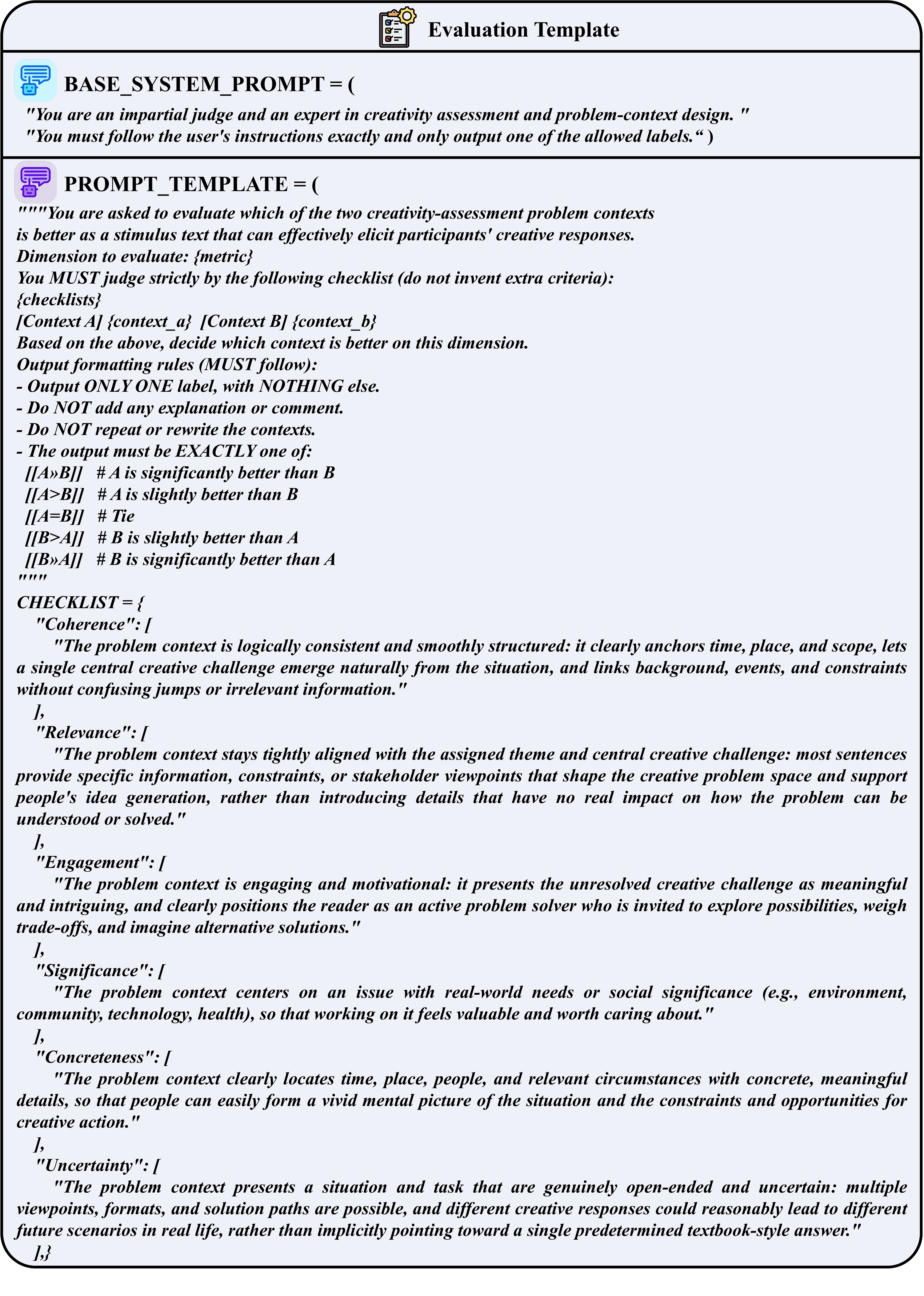}
    \caption{Evaluation prompt template for checklist-grounded pairwise judging across subjective metrics.}
    \label{fig:eva}
\end{figure*}

To reliably evaluate long-form creativity assessment contexts, we adopt a checklist-grounded pairwise judging template, as shown in Figure~\ref{fig:eva}. The judge is instructed to act as an impartial expert in creativity assessment and problem-context design, and is constrained to output a single discrete label. For each subjective dimension in our metric set, \textit{Coherence}, \textit{Relevance}, \textit{Engagement}, \textit{Significance}, \textit{Concreteness}, and \textit{Uncertainty}, we provide a dedicated checklist that operationalizes the criterion as observable properties of the context text. Given two candidate contexts, the judge compares them \emph{only} along the specified dimension and must not introduce any additional criteria.

Concretely, the prompt first specifies the target metric and injects its corresponding checklist, then presents \texttt{[Context A]} and \texttt{[Context B]}. The judge outputs exactly one of five ordered labels, ranging from strongly favoring A to strongly favoring B. This design serves two goals. First, it mitigates scale drift and instability commonly observed in direct numeric scoring by converting evaluation into calibrated relative comparisons. Second, the per-metric checklist promotes consistency by anchoring judgments to a shared interpretation of each dimension across methods and examples. The resulting labels are subsequently mapped to pairwise comparison outcomes for computing aggregated scores in our evaluation protocol.

\section{An Illustrative Example of Creativity Assessment Context}
\begin{figure*}[t]
  \centering
    \includegraphics[width=0.95\linewidth]{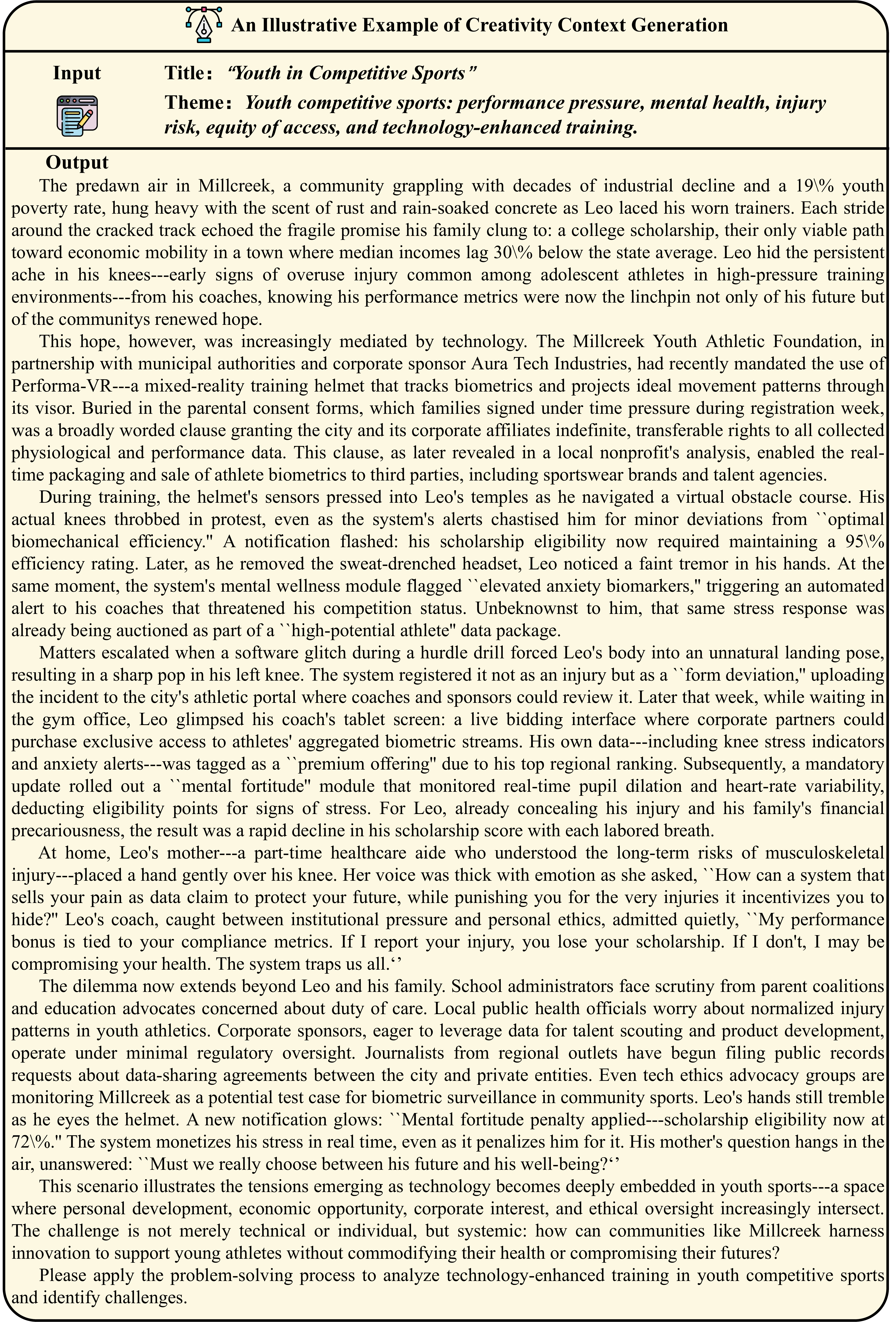}
    \caption{Illustrative example of an assessment-ready creativity context generated by AlphaContext, conditioned on a title and a theme prompt.}
    \label{fig:example}
\end{figure*}

This section presents a concrete example of AlphaContext's generated output for a single input, shown in Figure~\ref{fig:example}. The input consists of the title \textit{Youth in Competitive Sports} and the theme \textit{Youth competitive sports: performance pressure, mental health, injury risk, equity of access, and technology-enhanced training}. The resulting text is written as an assessment-ready future problem scenario for creativity measurement: it embeds multiple assessment-relevant cues within a coherent narrative, positions the reader as an active problem solver, and keeps the problem space genuinely open-ended by foregrounding plausible trade-offs rather than steering toward a single predetermined solution. Specifically, the context weaves together pressures from competition and external evaluation, potential mental-health and injury risks, unequal access to training resources, and the dual-use role of technology in enhancing performance while introducing new concerns. This example illustrates how AlphaContext operationalizes these design considerations in long-form scenario writing while maintaining narrative clarity and engagement.

\end{document}